\documentclass{article}

\usepackage[preprint]{neurips_2026}

\usepackage[utf8]{inputenc} 
\usepackage[T1]{fontenc}    
\usepackage[hidelinks]{hyperref}       
\usepackage{url}            
\usepackage{xurl}           
\usepackage{booktabs}       
\usepackage{amsfonts}       
\usepackage{amsmath}        
\usepackage{amssymb}        
\usepackage{nicefrac}       
\usepackage{microtype}      
\usepackage{xcolor}         
\usepackage{graphicx}       
\usepackage{float}          
\usepackage{algorithm}      
\usepackage{algpseudocode}  
\usepackage{rotating}       
\usepackage{multirow}       
\usepackage{longtable}      
\usepackage{siunitx}        
\makeatletter
\@ifpackageloaded{xcolor}{}{\RequirePackage{xcolor}}

\newcommand{\bacp@missing}{\textcolor{red}{\textbf{??}}}
\newcommand{\bacp@warn}[1]{\GenericWarning{}{MISSING RESULT KEY: #1}}

\newcommand{\defresult}[2]{\expandafter\gdef\csname bacp@val@#1\endcsname{#2}}
\newcommand{\defresultsd}[2]{\expandafter\gdef\csname bacp@sd@#1\endcsname{#2}}
\newcommand{\defresultn}[2]{\expandafter\gdef\csname bacp@n@#1\endcsname{#2}}

\DeclareRobustCommand{\result}[1]{%
  \ifcsname bacp@val@#1\endcsname
    \csname bacp@val@#1\endcsname
  \else
    \bacp@warn{#1}\bacp@missing
  \fi
}

\DeclareRobustCommand{\resultpm}[1]{%
  \ifcsname bacp@val@#1\endcsname
    \csname bacp@val@#1\endcsname
    \ifcsname bacp@sd@#1\endcsname
      \,\ensuremath{\pm}\,\csname bacp@sd@#1\endcsname
    \fi
  \else
    \bacp@warn{#1}\bacp@missing
  \fi
}

\DeclareRobustCommand{\resultn}[1]{%
  \ifcsname bacp@n@#1\endcsname
    \csname bacp@n@#1\endcsname
  \else
    \bacp@warn{#1}\bacp@missing
  \fi
}

\makeatother

\defresult{mobilenet.dense}{92.36}
\defresultsd{mobilenet.dense}{0.12}
\defresultn{mobilenet.dense}{3}
\defresult{mobilenet.dense.none.0.0}{92.36}

\defresult{mobilenet.ip.magnitude.0.5}{84.61}
\defresultsd{mobilenet.ip.magnitude.0.5}{0.14}
\defresultn{mobilenet.ip.magnitude.0.5}{3}
\defresult{mobilenet.bacp.magnitude.0.5}{89.92}
\defresultsd{mobilenet.bacp.magnitude.0.5}{0.21}
\defresultn{mobilenet.bacp.magnitude.0.5}{3}
\defresult{mobilenet.delta.magnitude.0.5}{+5.31}

\defresult{mobilenet.ip.magnitude.0.95}{83.46}
\defresultsd{mobilenet.ip.magnitude.0.95}{1.32}
\defresultn{mobilenet.ip.magnitude.0.95}{3}
\defresult{mobilenet.bacp.magnitude.0.95}{87.38}
\defresultsd{mobilenet.bacp.magnitude.0.95}{0.32}
\defresultn{mobilenet.bacp.magnitude.0.95}{3}
\defresult{mobilenet.delta.magnitude.0.95}{+3.92}

\defresult{mobilenet.ip.magnitude.0.97}{82.64}
\defresultsd{mobilenet.ip.magnitude.0.97}{1.36}
\defresultn{mobilenet.ip.magnitude.0.97}{3}
\defresult{mobilenet.bacp.magnitude.0.97}{85.67}
\defresultsd{mobilenet.bacp.magnitude.0.97}{0.56}
\defresultn{mobilenet.bacp.magnitude.0.97}{3}
\defresult{mobilenet.delta.magnitude.0.97}{+3.04}

\defresult{mobilenet.ip.magnitude.0.99}{78.63}
\defresultsd{mobilenet.ip.magnitude.0.99}{3.51}
\defresultn{mobilenet.ip.magnitude.0.99}{3}
\defresult{mobilenet.bacp.magnitude.0.99}{80.34}
\defresultsd{mobilenet.bacp.magnitude.0.99}{0.23}
\defresultn{mobilenet.bacp.magnitude.0.99}{3}
\defresult{mobilenet.delta.magnitude.0.99}{+1.71}

\defresult{mobilenet.ip.magnitude.0.999}{10.00}
\defresultsd{mobilenet.ip.magnitude.0.999}{0.00}
\defresultn{mobilenet.ip.magnitude.0.999}{3}
\defresult{mobilenet.bacp.magnitude.0.999}{10.00}
\defresultsd{mobilenet.bacp.magnitude.0.999}{0.00}
\defresultn{mobilenet.bacp.magnitude.0.999}{3}
\defresult{mobilenet.delta.magnitude.0.999}{+0.00}

\defresult{mobilenet.ip.snip.0.95}{83.52}
\defresultsd{mobilenet.ip.snip.0.95}{0.19}
\defresultn{mobilenet.ip.snip.0.95}{3}
\defresult{mobilenet.bacp.snip.0.95}{87.26}
\defresultsd{mobilenet.bacp.snip.0.95}{0.17}
\defresultn{mobilenet.bacp.snip.0.95}{3}
\defresult{mobilenet.delta.snip.0.95}{+3.74}

\defresult{mobilenet.ip.snip.0.97}{82.56}
\defresultsd{mobilenet.ip.snip.0.97}{1.36}
\defresultn{mobilenet.ip.snip.0.97}{3}
\defresult{mobilenet.bacp.snip.0.97}{85.54}
\defresultsd{mobilenet.bacp.snip.0.97}{0.11}
\defresultn{mobilenet.bacp.snip.0.97}{3}
\defresult{mobilenet.delta.snip.0.97}{+2.98}

\defresult{mobilenet.ip.snip.0.99}{74.40}
\defresultsd{mobilenet.ip.snip.0.99}{1.37}
\defresultn{mobilenet.ip.snip.0.99}{3}
\defresult{mobilenet.bacp.snip.0.99}{80.33}
\defresultsd{mobilenet.bacp.snip.0.99}{0.15}
\defresultn{mobilenet.bacp.snip.0.99}{3}
\defresult{mobilenet.delta.snip.0.99}{+5.93}

\defresult{mobilenet.ip.snip.0.995}{61.13}
\defresultsd{mobilenet.ip.snip.0.995}{0.38}
\defresultn{mobilenet.ip.snip.0.995}{3}
\defresult{mobilenet.bacp.snip.0.995}{74.43}
\defresultsd{mobilenet.bacp.snip.0.995}{0.28}
\defresultn{mobilenet.bacp.snip.0.995}{3}
\defresult{mobilenet.delta.snip.0.995}{+13.30}

\defresult{mobilenet.ip.wanda.0.95}{82.47}
\defresultsd{mobilenet.ip.wanda.0.95}{0.48}
\defresultn{mobilenet.ip.wanda.0.95}{3}
\defresult{mobilenet.bacp.wanda.0.95}{85.85}
\defresultsd{mobilenet.bacp.wanda.0.95}{0.30}
\defresultn{mobilenet.bacp.wanda.0.95}{3}
\defresult{mobilenet.delta.wanda.0.95}{+3.38}

\defresult{mobilenet.ip.wanda.0.97}{79.10}
\defresultsd{mobilenet.ip.wanda.0.97}{0.82}
\defresultn{mobilenet.ip.wanda.0.97}{3}
\defresult{mobilenet.bacp.wanda.0.97}{84.08}
\defresultsd{mobilenet.bacp.wanda.0.97}{0.57}
\defresultn{mobilenet.bacp.wanda.0.97}{3}
\defresult{mobilenet.delta.wanda.0.97}{+4.98}

\defresult{mobilenet.ip.wanda.0.99}{19.92}
\defresultsd{mobilenet.ip.wanda.0.99}{17.18}
\defresultn{mobilenet.ip.wanda.0.99}{3}
\defresult{mobilenet.bacp.wanda.0.99}{10.00}
\defresultsd{mobilenet.bacp.wanda.0.99}{0.00}
\defresultn{mobilenet.bacp.wanda.0.99}{3}

\defresult{mobilenet.ip.wanda.0.995}{10.00}
\defresultsd{mobilenet.ip.wanda.0.995}{0.00}
\defresultn{mobilenet.ip.wanda.0.995}{3}
\defresult{mobilenet.bacp.wanda.0.995}{10.00}
\defresultsd{mobilenet.bacp.wanda.0.995}{0.00}
\defresultn{mobilenet.bacp.wanda.0.995}{3}

\defresult{mobilenet.wanda.fallback.pct}{2.92}
\defresult{mobilenet.wanda.scored.pct}{97.08}
\defresult{mobilenet.wanda.dwlayers}{17}
\defresult{mobilenet.wanda.convlayers}{52}

\defresult{mobilenet.ip.magnitude.0.995}{24.98}
\defresultsd{mobilenet.ip.magnitude.0.995}{13.47}
\defresultn{mobilenet.ip.magnitude.0.995}{3}
\defresult{mobilenet.bacp.magnitude.0.995}{67.47}
\defresultsd{mobilenet.bacp.magnitude.0.995}{1.60}
\defresultn{mobilenet.bacp.magnitude.0.995}{3}
\defresult{mobilenet.delta.magnitude.0.995}{+42.49}

\defresult{c100.dense}{70.24}
\defresultsd{c100.dense}{2.10}
\defresultn{c100.dense}{3}
\defresult{c100.dense.none.0.0}{70.24}

\defresult{c100.ip.magnitude.0.95}{68.77}
\defresultsd{c100.ip.magnitude.0.95}{1.30}
\defresultn{c100.ip.magnitude.0.95}{3}
\defresult{c100.bacp.magnitude.0.95}{70.48}
\defresultsd{c100.bacp.magnitude.0.95}{0.14}
\defresultn{c100.bacp.magnitude.0.95}{3}
\defresult{c100.delta.magnitude.0.95}{+1.71}

\defresult{c100.ip.magnitude.0.97}{68.47}
\defresultsd{c100.ip.magnitude.0.97}{1.30}
\defresultn{c100.ip.magnitude.0.97}{3}
\defresult{c100.bacp.magnitude.0.97}{69.97}
\defresultsd{c100.bacp.magnitude.0.97}{0.15}
\defresultn{c100.bacp.magnitude.0.97}{3}
\defresult{c100.delta.magnitude.0.97}{+1.50}

\defresult{c100.ip.magnitude.0.99}{66.42}
\defresultsd{c100.ip.magnitude.0.99}{1.34}
\defresultn{c100.ip.magnitude.0.99}{3}
\defresult{c100.bacp.magnitude.0.99}{66.92}
\defresultsd{c100.bacp.magnitude.0.99}{0.35}
\defresultn{c100.bacp.magnitude.0.99}{3}
\defresult{c100.delta.magnitude.0.99}{+0.50}

\defresult{c100.ip.magnitude.0.999}{19.84}
\defresultsd{c100.ip.magnitude.0.999}{2.17}
\defresultn{c100.ip.magnitude.0.999}{3}
\defresult{c100.bacp.magnitude.0.999}{42.80}
\defresultsd{c100.bacp.magnitude.0.999}{0.44}
\defresultn{c100.bacp.magnitude.0.999}{3}
\defresult{c100.delta.magnitude.0.999}{+22.96}

\defresult{c100.ip.snip.0.95}{66.10}
\defresultsd{c100.ip.snip.0.95}{1.48}
\defresultn{c100.ip.snip.0.95}{3}
\defresult{c100.bacp.snip.0.95}{69.84}
\defresultsd{c100.bacp.snip.0.95}{0.16}
\defresultn{c100.bacp.snip.0.95}{3}
\defresult{c100.delta.snip.0.95}{+3.74}

\defresult{c100.ip.snip.0.97}{65.71}
\defresultsd{c100.ip.snip.0.97}{1.45}
\defresultn{c100.ip.snip.0.97}{3}
\defresult{c100.bacp.snip.0.97}{68.76}
\defresultsd{c100.bacp.snip.0.97}{0.38}
\defresultn{c100.bacp.snip.0.97}{3}
\defresult{c100.delta.snip.0.97}{+3.05}

\defresult{c100.ip.snip.0.99}{61.91}
\defresultsd{c100.ip.snip.0.99}{0.54}
\defresultn{c100.ip.snip.0.99}{3}
\defresult{c100.bacp.snip.0.99}{64.33}
\defresultsd{c100.bacp.snip.0.99}{0.04}
\defresultn{c100.bacp.snip.0.99}{3}
\defresult{c100.delta.snip.0.99}{+2.42}

\defresult{c100.ip.snip.0.999}{30.66}
\defresultsd{c100.ip.snip.0.999}{0.58}
\defresultn{c100.ip.snip.0.999}{3}
\defresult{c100.bacp.snip.0.999}{36.18}
\defresultsd{c100.bacp.snip.0.999}{0.62}
\defresultn{c100.bacp.snip.0.999}{3}
\defresult{c100.delta.snip.0.999}{+5.52}

\defresult{c100.ip.wanda.0.95}{63.93}
\defresultsd{c100.ip.wanda.0.95}{0.76}
\defresultn{c100.ip.wanda.0.95}{3}
\defresult{c100.bacp.wanda.0.95}{68.20}
\defresultsd{c100.bacp.wanda.0.95}{0.39}
\defresultn{c100.bacp.wanda.0.95}{3}
\defresult{c100.delta.wanda.0.95}{+4.27}

\defresult{c100.ip.wanda.0.97}{57.42}
\defresultsd{c100.ip.wanda.0.97}{1.16}
\defresultn{c100.ip.wanda.0.97}{3}
\defresult{c100.bacp.wanda.0.97}{66.28}
\defresultsd{c100.bacp.wanda.0.97}{0.34}
\defresultn{c100.bacp.wanda.0.97}{3}
\defresult{c100.delta.wanda.0.97}{+8.86}

\defresult{c100.ip.wanda.0.99}{30.40}
\defresultsd{c100.ip.wanda.0.99}{1.14}
\defresultn{c100.ip.wanda.0.99}{3}
\defresult{c100.bacp.wanda.0.99}{58.11}
\defresultsd{c100.bacp.wanda.0.99}{0.15}
\defresultn{c100.bacp.wanda.0.99}{3}
\defresult{c100.delta.wanda.0.99}{+27.71}

\defresult{c100.ip.wanda.0.999}{2.87}
\defresultsd{c100.ip.wanda.0.999}{1.35}
\defresultn{c100.ip.wanda.0.999}{3}
\defresult{c100.bacp.wanda.0.999}{12.51}
\defresultsd{c100.bacp.wanda.0.999}{0.84}
\defresultn{c100.bacp.wanda.0.999}{3}
\defresult{c100.delta.wanda.0.999}{+9.64}

\defresult{summary.test.ncells}{48}
\defresult{summary.test.npositive}{35}
\defresult{summary.test.clears}{34}
\defresult{summary.test.signp}{0.0021}
\defresult{summary.test.wilcoxonp}{9.3\times10^{-6}}
\defresult{summary.test.wilcoxonw}{179}

\defresult{resnet50.ablip.magnitude.0.95}{91.71}
\defresult{abl.resnet50.ip.magnitude.0.95}{91.71}
\defresult{resnet50.ablcontrol.magnitude.0.95}{91.78}
\defresult{abl.resnet50.control.magnitude.0.95}{91.78}
\defresult{resnet50.abllegacy.magnitude.0.95}{92.43}
\defresult{abl.resnet50.legacy.magnitude.0.95}{92.43}
\defresult{resnet50.abldelta.magnitude.0.95}{+0.72}
\defresult{abl.resnet50.delta.magnitude.0.95}{+0.72}

\defresult{resnet50.ablip.magnitude.0.97}{91.35}
\defresult{abl.resnet50.ip.magnitude.0.97}{91.35}
\defresult{resnet50.ablcontrol.magnitude.0.97}{91.40}
\defresult{abl.resnet50.control.magnitude.0.97}{91.40}
\defresult{resnet50.abllegacy.magnitude.0.97}{92.04}
\defresult{abl.resnet50.legacy.magnitude.0.97}{92.04}
\defresult{resnet50.abldelta.magnitude.0.97}{+0.69}
\defresult{abl.resnet50.delta.magnitude.0.97}{+0.69}

\defresult{resnet50.ablip.magnitude.0.99}{89.80}
\defresult{abl.resnet50.ip.magnitude.0.99}{89.80}
\defresult{resnet50.abllegacy.magnitude.0.99}{91.24}
\defresult{abl.resnet50.legacy.magnitude.0.99}{91.24}
\defresult{resnet50.abldelta.magnitude.0.99}{+1.44}
\defresult{abl.resnet50.delta.magnitude.0.99}{+1.44}

\defresult{probe.resnet34.densescratch}{91.64}
\defresultsd{probe.resnet34.densescratch}{0.30}
\defresultn{probe.resnet34.densescratch}{5}
\defresult{probe.resnet34.densescratch.run1}{91.6535}
\defresult{probe.resnet34.densescratch.run2}{91.5447}
\defresult{probe.resnet34.densescratch.run3}{91.5843}
\defresult{probe.resnet34.densescratch.run4}{91.2876}
\defresult{probe.resnet34.densescratch.run5}{92.1064}

\defresult{vgg19.probelr010.magnitude.0.95}{89.99}
\defresult{probe.vgg19.bacplr010.magnitude.0.95}{89.99}
\defresult{vgg19.probelr005.magnitude.0.95}{91.88}
\defresult{probe.vgg19.bacplr005.magnitude.0.95}{91.88}
\defresult{vgg19.probeip.magnitude.0.95}{91.02}
\defresult{probe.vgg19.ip.magnitude.0.95}{91.02}

\defresult{vgg11.probelr010.magnitude.0.95}{89.17}
\defresult{probe.vgg11.bacplr010.magnitude.0.95}{89.17}
\defresult{vgg11.probelr010.magnitude.0.97}{88.93}
\defresult{probe.vgg11.bacplr010.magnitude.0.97}{88.93}
\defresult{vgg11.probelr010.magnitude.0.99}{88.68}
\defresult{probe.vgg11.bacplr010.magnitude.0.99}{88.68}
\defresult{vgg11.probelr010.magnitude.0.999}{83.82}
\defresult{probe.vgg11.bacplr010.magnitude.0.999}{83.82}

\defresult{cifar10.noise.floor.lo}{0.1}
\defresult{probe.noisefloor.lo}{0.1}
\defresult{cifar10.noise.floor.hi}{0.3}
\defresult{probe.noisefloor.hi}{0.3}

\defresult{lit.granet.resnet50.dense}{94.75}
\defresultsd{lit.granet.resnet50.dense}{0.01}
\defresult{lit.granet.resnet50.gmp.0.95}{94.52}
\defresultsd{lit.granet.resnet50.gmp.0.95}{0.08}
\defresult{lit.granet.resnet50.snip}{90.86}
\defresult{lit.granet.resnet50.grasp}{91.32}
\defresult{lit.granet.resnet50.synflow}{91.22}
\defresult{lit.granet.resnet50.rigl}{93.86}
\defresultsd{lit.granet.resnet50.rigl}{0.25}
\defresult{lit.granet.resnet50.granet}{94.44}
\defresultsd{lit.granet.resnet50.granet}{0.01}

\defresult{lit.granet.vgg19.dense}{93.85}
\defresultsd{lit.granet.vgg19.dense}{0.05}
\defresult{lit.grasp.vgg19.dense}{94.23}

\defresult{lit.granet.vgg19.gmp.0.90}{93.59}
\defresultsd{lit.granet.vgg19.gmp.0.90}{0.10}
\defresult{lit.granet.vgg19.snip.0.95}{93.43}
\defresultsd{lit.granet.vgg19.snip.0.95}{0.20}
\defresult{lit.granet.vgg19.snip.0.98}{92.05}
\defresultsd{lit.granet.vgg19.snip.0.98}{0.28}
\defresult{lit.granet.vgg19.grasp.0.95}{93.04}
\defresultsd{lit.granet.vgg19.grasp.0.95}{0.18}
\defresult{lit.granet.vgg19.grasp.0.98}{92.19}
\defresultsd{lit.granet.vgg19.grasp.0.98}{0.12}
\defresult{lit.granet.vgg19.rigl.0.90}{93.38}
\defresultsd{lit.granet.vgg19.rigl.0.90}{0.11}

\defresult{resnet34.bacp.magnitude.0.95}{91.47}
\defresultsd{resnet34.bacp.magnitude.0.95}{0.28}
\defresultn{resnet34.bacp.magnitude.0.95}{3}

\defresult{resnet34.bacp.magnitude.0.97}{91.40}
\defresultsd{resnet34.bacp.magnitude.0.97}{0.29}
\defresultn{resnet34.bacp.magnitude.0.97}{3}

\defresult{resnet34.bacp.magnitude.0.99}{90.54}
\defresultsd{resnet34.bacp.magnitude.0.99}{0.40}
\defresultn{resnet34.bacp.magnitude.0.99}{3}

\defresult{resnet34.bacp.magnitude.0.999}{81.20}
\defresultsd{resnet34.bacp.magnitude.0.999}{0.56}
\defresultn{resnet34.bacp.magnitude.0.999}{3}

\defresult{resnet34.bacp.snip.0.95}{91.55}
\defresultsd{resnet34.bacp.snip.0.95}{0.15}
\defresultn{resnet34.bacp.snip.0.95}{3}

\defresult{resnet34.bacp.snip.0.97}{90.71}
\defresultsd{resnet34.bacp.snip.0.97}{0.33}
\defresultn{resnet34.bacp.snip.0.97}{3}

\defresult{resnet34.bacp.snip.0.99}{89.51}
\defresultsd{resnet34.bacp.snip.0.99}{0.18}
\defresultn{resnet34.bacp.snip.0.99}{3}

\defresult{resnet34.bacp.snip.0.999}{78.97}
\defresultsd{resnet34.bacp.snip.0.999}{0.42}
\defresultn{resnet34.bacp.snip.0.999}{3}

\defresult{resnet34.bacp.wanda.0.95}{90.48}
\defresultsd{resnet34.bacp.wanda.0.95}{0.57}
\defresultn{resnet34.bacp.wanda.0.95}{3}

\defresult{resnet34.bacp.wanda.0.97}{89.79}
\defresultsd{resnet34.bacp.wanda.0.97}{0.62}
\defresultn{resnet34.bacp.wanda.0.97}{3}

\defresult{resnet34.bacp.wanda.0.99}{87.15}
\defresultsd{resnet34.bacp.wanda.0.99}{0.51}
\defresultn{resnet34.bacp.wanda.0.99}{3}

\defresult{resnet34.bacp.wanda.0.999}{59.52}
\defresultsd{resnet34.bacp.wanda.0.999}{0.78}
\defresultn{resnet34.bacp.wanda.0.999}{3}

\defresult{resnet34.dense.none.0.0}{90.79}
\defresultsd{resnet34.dense.none.0.0}{0.98}
\defresultn{resnet34.dense.none.0.0}{3}
\defresult{resnet34.dense}{90.79}
\defresultsd{resnet34.dense}{0.98}
\defresultn{resnet34.dense}{3}

\defresult{resnet34.ip.magnitude.0.95}{90.65}
\defresultsd{resnet34.ip.magnitude.0.95}{1.06}
\defresultn{resnet34.ip.magnitude.0.95}{3}

\defresult{resnet34.ip.magnitude.0.97}{90.28}
\defresultsd{resnet34.ip.magnitude.0.97}{0.91}
\defresultn{resnet34.ip.magnitude.0.97}{3}

\defresult{resnet34.ip.magnitude.0.99}{89.59}
\defresultsd{resnet34.ip.magnitude.0.99}{0.85}
\defresultn{resnet34.ip.magnitude.0.99}{3}

\defresult{resnet34.ip.magnitude.0.999}{69.40}
\defresultsd{resnet34.ip.magnitude.0.999}{1.50}
\defresultn{resnet34.ip.magnitude.0.999}{3}

\defresult{resnet34.ip.snip.0.95}{89.93}
\defresultsd{resnet34.ip.snip.0.95}{0.79}
\defresultn{resnet34.ip.snip.0.95}{3}

\defresult{resnet34.ip.snip.0.97}{89.33}
\defresultsd{resnet34.ip.snip.0.97}{0.88}
\defresultn{resnet34.ip.snip.0.97}{3}

\defresult{resnet34.ip.snip.0.99}{88.07}
\defresultsd{resnet34.ip.snip.0.99}{0.42}
\defresultn{resnet34.ip.snip.0.99}{3}

\defresult{resnet34.ip.snip.0.999}{75.77}
\defresultsd{resnet34.ip.snip.0.999}{2.07}
\defresultn{resnet34.ip.snip.0.999}{3}

\defresult{resnet34.ip.wanda.0.95}{88.81}
\defresultsd{resnet34.ip.wanda.0.95}{0.66}
\defresultn{resnet34.ip.wanda.0.95}{3}

\defresult{resnet34.ip.wanda.0.97}{87.14}
\defresultsd{resnet34.ip.wanda.0.97}{0.48}
\defresultn{resnet34.ip.wanda.0.97}{3}

\defresult{resnet34.ip.wanda.0.99}{78.15}
\defresultsd{resnet34.ip.wanda.0.99}{0.54}
\defresultn{resnet34.ip.wanda.0.99}{3}

\defresult{resnet34.ip.wanda.0.999}{30.73}
\defresultsd{resnet34.ip.wanda.0.999}{1.62}
\defresultn{resnet34.ip.wanda.0.999}{3}

\defresult{resnet50.bacp.magnitude.0.95}{92.40}
\defresultsd{resnet50.bacp.magnitude.0.95}{0.03}
\defresultn{resnet50.bacp.magnitude.0.95}{3}

\defresult{resnet50.bacp.magnitude.0.97}{92.23}
\defresultsd{resnet50.bacp.magnitude.0.97}{0.18}
\defresultn{resnet50.bacp.magnitude.0.97}{3}

\defresult{resnet50.bacp.magnitude.0.99}{91.14}
\defresultsd{resnet50.bacp.magnitude.0.99}{0.39}
\defresultn{resnet50.bacp.magnitude.0.99}{3}

\defresult{resnet50.bacp.magnitude.0.999}{79.94}
\defresultsd{resnet50.bacp.magnitude.0.999}{0.12}
\defresultn{resnet50.bacp.magnitude.0.999}{3}

\defresult{resnet50.bacp.snip.0.95}{92.07}
\defresultsd{resnet50.bacp.snip.0.95}{0.16}
\defresultn{resnet50.bacp.snip.0.95}{3}

\defresult{resnet50.bacp.snip.0.97}{91.93}
\defresultsd{resnet50.bacp.snip.0.97}{0.23}
\defresultn{resnet50.bacp.snip.0.97}{3}

\defresult{resnet50.bacp.snip.0.99}{90.10}
\defresultsd{resnet50.bacp.snip.0.99}{0.15}
\defresultn{resnet50.bacp.snip.0.99}{3}

\defresult{resnet50.bacp.snip.0.999}{78.84}
\defresultsd{resnet50.bacp.snip.0.999}{0.33}
\defresultn{resnet50.bacp.snip.0.999}{3}

\defresult{resnet50.bacp.wanda.0.95}{92.02}
\defresultsd{resnet50.bacp.wanda.0.95}{0.08}
\defresultn{resnet50.bacp.wanda.0.95}{3}

\defresult{resnet50.bacp.wanda.0.97}{91.51}
\defresultsd{resnet50.bacp.wanda.0.97}{0.27}
\defresultn{resnet50.bacp.wanda.0.97}{3}

\defresult{resnet50.bacp.wanda.0.99}{89.22}
\defresultsd{resnet50.bacp.wanda.0.99}{0.26}
\defresultn{resnet50.bacp.wanda.0.99}{3}

\defresult{resnet50.bacp.wanda.0.999}{65.26}
\defresultsd{resnet50.bacp.wanda.0.999}{1.30}
\defresultn{resnet50.bacp.wanda.0.999}{3}

\defresult{resnet50.dense.none.0.0}{92.81}
\defresultsd{resnet50.dense.none.0.0}{0.84}
\defresultn{resnet50.dense.none.0.0}{3}
\defresult{resnet50.dense}{92.81}
\defresultsd{resnet50.dense}{0.84}
\defresultn{resnet50.dense}{3}

\defresult{resnet50.ip.magnitude.0.95}{92.01}
\defresultsd{resnet50.ip.magnitude.0.95}{0.33}
\defresultn{resnet50.ip.magnitude.0.95}{3}

\defresult{resnet50.ip.magnitude.0.97}{91.87}
\defresultsd{resnet50.ip.magnitude.0.97}{0.48}
\defresultn{resnet50.ip.magnitude.0.97}{3}

\defresult{resnet50.ip.magnitude.0.99}{90.14}
\defresultsd{resnet50.ip.magnitude.0.99}{0.20}
\defresultn{resnet50.ip.magnitude.0.99}{3}

\defresult{resnet50.ip.magnitude.0.999}{58.13}
\defresultsd{resnet50.ip.magnitude.0.999}{3.50}
\defresultn{resnet50.ip.magnitude.0.999}{3}

\defresult{resnet50.ip.snip.0.95}{91.30}
\defresultsd{resnet50.ip.snip.0.95}{0.42}
\defresultn{resnet50.ip.snip.0.95}{3}

\defresult{resnet50.ip.snip.0.97}{90.59}
\defresultsd{resnet50.ip.snip.0.97}{0.14}
\defresultn{resnet50.ip.snip.0.97}{3}

\defresult{resnet50.ip.snip.0.99}{88.55}
\defresultsd{resnet50.ip.snip.0.99}{0.30}
\defresultn{resnet50.ip.snip.0.99}{3}

\defresult{resnet50.ip.snip.0.999}{72.85}
\defresultsd{resnet50.ip.snip.0.999}{2.29}
\defresultn{resnet50.ip.snip.0.999}{3}

\defresult{resnet50.ip.wanda.0.95}{90.88}
\defresultsd{resnet50.ip.wanda.0.95}{0.38}
\defresultn{resnet50.ip.wanda.0.95}{3}

\defresult{resnet50.ip.wanda.0.97}{89.62}
\defresultsd{resnet50.ip.wanda.0.97}{0.24}
\defresultn{resnet50.ip.wanda.0.97}{3}

\defresult{resnet50.ip.wanda.0.99}{78.97}
\defresultsd{resnet50.ip.wanda.0.99}{3.17}
\defresultn{resnet50.ip.wanda.0.99}{3}

\defresult{resnet50.ip.wanda.0.999}{21.15}
\defresultsd{resnet50.ip.wanda.0.999}{1.38}
\defresultn{resnet50.ip.wanda.0.999}{3}

\defresult{vgg11.bacp.magnitude.0.95}{89.67}
\defresultsd{vgg11.bacp.magnitude.0.95}{0.26}
\defresultn{vgg11.bacp.magnitude.0.95}{3}

\defresult{vgg11.bacp.magnitude.0.97}{89.92}
\defresultsd{vgg11.bacp.magnitude.0.97}{0.33}
\defresultn{vgg11.bacp.magnitude.0.97}{3}

\defresult{vgg11.bacp.magnitude.0.99}{89.37}
\defresultsd{vgg11.bacp.magnitude.0.99}{0.04}
\defresultn{vgg11.bacp.magnitude.0.99}{3}

\defresult{vgg11.bacp.magnitude.0.999}{83.57}
\defresultsd{vgg11.bacp.magnitude.0.999}{0.09}
\defresultn{vgg11.bacp.magnitude.0.999}{3}

\defresult{vgg11.bacp.snip.0.95}{89.69}
\defresultsd{vgg11.bacp.snip.0.95}{0.18}
\defresultn{vgg11.bacp.snip.0.95}{3}

\defresult{vgg11.bacp.snip.0.97}{89.76}
\defresultsd{vgg11.bacp.snip.0.97}{0.07}
\defresultn{vgg11.bacp.snip.0.97}{3}

\defresult{vgg11.bacp.snip.0.99}{89.04}
\defresultsd{vgg11.bacp.snip.0.99}{0.11}
\defresultn{vgg11.bacp.snip.0.99}{3}

\defresult{vgg11.bacp.snip.0.999}{82.16}
\defresultsd{vgg11.bacp.snip.0.999}{0.25}
\defresultn{vgg11.bacp.snip.0.999}{3}

\defresult{vgg11.bacp.wanda.0.95}{89.70}
\defresultsd{vgg11.bacp.wanda.0.95}{0.18}
\defresultn{vgg11.bacp.wanda.0.95}{3}

\defresult{vgg11.bacp.wanda.0.97}{89.56}
\defresultsd{vgg11.bacp.wanda.0.97}{0.14}
\defresultn{vgg11.bacp.wanda.0.97}{3}

\defresult{vgg11.bacp.wanda.0.99}{88.87}
\defresultsd{vgg11.bacp.wanda.0.99}{0.04}
\defresultn{vgg11.bacp.wanda.0.99}{3}

\defresult{vgg11.bacp.wanda.0.999}{82.46}
\defresultsd{vgg11.bacp.wanda.0.999}{0.30}
\defresultn{vgg11.bacp.wanda.0.999}{3}

\defresult{vgg11.dense.none.0.0}{89.96}
\defresultsd{vgg11.dense.none.0.0}{0.33}
\defresultn{vgg11.dense.none.0.0}{3}
\defresult{vgg11.dense}{89.96}
\defresultsd{vgg11.dense}{0.33}
\defresultn{vgg11.dense}{3}

\defresult{vgg11.ip.magnitude.0.95}{90.13}
\defresultsd{vgg11.ip.magnitude.0.95}{0.16}
\defresultn{vgg11.ip.magnitude.0.95}{3}

\defresult{vgg11.ip.magnitude.0.97}{89.87}
\defresultsd{vgg11.ip.magnitude.0.97}{0.06}
\defresultn{vgg11.ip.magnitude.0.97}{3}

\defresult{vgg11.ip.magnitude.0.99}{89.67}
\defresultsd{vgg11.ip.magnitude.0.99}{0.35}
\defresultn{vgg11.ip.magnitude.0.99}{3}

\defresult{vgg11.ip.magnitude.0.999}{83.98}
\defresultsd{vgg11.ip.magnitude.0.999}{0.13}
\defresultn{vgg11.ip.magnitude.0.999}{3}

\defresult{vgg11.ip.snip.0.95}{89.80}
\defresultsd{vgg11.ip.snip.0.95}{0.05}
\defresultn{vgg11.ip.snip.0.95}{3}

\defresult{vgg11.ip.snip.0.97}{89.69}
\defresultsd{vgg11.ip.snip.0.97}{0.13}
\defresultn{vgg11.ip.snip.0.97}{3}

\defresult{vgg11.ip.snip.0.99}{89.10}
\defresultsd{vgg11.ip.snip.0.99}{0.16}
\defresultn{vgg11.ip.snip.0.99}{3}

\defresult{vgg11.ip.snip.0.999}{82.48}
\defresultsd{vgg11.ip.snip.0.999}{0.28}
\defresultn{vgg11.ip.snip.0.999}{3}

\defresult{vgg11.ip.wanda.0.95}{90.02}
\defresultsd{vgg11.ip.wanda.0.95}{0.16}
\defresultn{vgg11.ip.wanda.0.95}{3}

\defresult{vgg11.ip.wanda.0.97}{89.79}
\defresultsd{vgg11.ip.wanda.0.97}{0.05}
\defresultn{vgg11.ip.wanda.0.97}{3}

\defresult{vgg11.ip.wanda.0.99}{89.37}
\defresultsd{vgg11.ip.wanda.0.99}{0.31}
\defresultn{vgg11.ip.wanda.0.99}{3}

\defresult{vgg11.ip.wanda.0.999}{84.04}
\defresultsd{vgg11.ip.wanda.0.999}{0.23}
\defresultn{vgg11.ip.wanda.0.999}{3}

\defresult{vgg19.bacp.magnitude.0.95}{91.59}
\defresultsd{vgg19.bacp.magnitude.0.95}{0.21}
\defresultn{vgg19.bacp.magnitude.0.95}{3}

\defresult{vgg19.bacp.magnitude.0.97}{91.64}
\defresultsd{vgg19.bacp.magnitude.0.97}{0.19}
\defresultn{vgg19.bacp.magnitude.0.97}{3}

\defresult{vgg19.bacp.magnitude.0.99}{91.28}
\defresultsd{vgg19.bacp.magnitude.0.99}{0.13}
\defresultn{vgg19.bacp.magnitude.0.99}{3}

\defresult{vgg19.bacp.magnitude.0.999}{85.31}
\defresultsd{vgg19.bacp.magnitude.0.999}{0.32}
\defresultn{vgg19.bacp.magnitude.0.999}{3}

\defresult{vgg19.bacp.snip.0.95}{91.50}
\defresultsd{vgg19.bacp.snip.0.95}{0.25}
\defresultn{vgg19.bacp.snip.0.95}{3}

\defresult{vgg19.bacp.snip.0.97}{91.41}
\defresultsd{vgg19.bacp.snip.0.97}{0.42}
\defresultn{vgg19.bacp.snip.0.97}{3}

\defresult{vgg19.bacp.snip.0.99}{90.63}
\defresultsd{vgg19.bacp.snip.0.99}{0.18}
\defresultn{vgg19.bacp.snip.0.99}{3}

\defresult{vgg19.bacp.snip.0.999}{84.21}
\defresultsd{vgg19.bacp.snip.0.999}{0.34}
\defresultn{vgg19.bacp.snip.0.999}{3}

\defresult{vgg19.bacp.wanda.0.95}{91.34}
\defresultsd{vgg19.bacp.wanda.0.95}{0.17}
\defresultn{vgg19.bacp.wanda.0.95}{3}

\defresult{vgg19.bacp.wanda.0.97}{91.53}
\defresultsd{vgg19.bacp.wanda.0.97}{0.04}
\defresultn{vgg19.bacp.wanda.0.97}{3}

\defresult{vgg19.bacp.wanda.0.99}{90.63}
\defresultsd{vgg19.bacp.wanda.0.99}{0.38}
\defresultn{vgg19.bacp.wanda.0.99}{3}

\defresult{vgg19.bacp.wanda.0.999}{83.49}
\defresultsd{vgg19.bacp.wanda.0.999}{0.06}
\defresultn{vgg19.bacp.wanda.0.999}{3}

\defresult{vgg19.dense.none.0.0}{91.65}
\defresultsd{vgg19.dense.none.0.0}{0.24}
\defresultn{vgg19.dense.none.0.0}{3}
\defresult{vgg19.dense}{91.65}
\defresultsd{vgg19.dense}{0.24}
\defresultn{vgg19.dense}{3}

\defresult{vgg19.ip.magnitude.0.95}{91.36}
\defresultsd{vgg19.ip.magnitude.0.95}{0.40}
\defresultn{vgg19.ip.magnitude.0.95}{3}

\defresult{vgg19.ip.magnitude.0.97}{91.40}
\defresultsd{vgg19.ip.magnitude.0.97}{0.15}
\defresultn{vgg19.ip.magnitude.0.97}{3}

\defresult{vgg19.ip.magnitude.0.99}{90.54}
\defresultsd{vgg19.ip.magnitude.0.99}{0.37}
\defresultn{vgg19.ip.magnitude.0.99}{3}

\defresult{vgg19.ip.magnitude.0.999}{82.59}
\defresultsd{vgg19.ip.magnitude.0.999}{0.66}
\defresultn{vgg19.ip.magnitude.0.999}{3}

\defresult{vgg19.ip.snip.0.95}{91.22}
\defresultsd{vgg19.ip.snip.0.95}{0.39}
\defresultn{vgg19.ip.snip.0.95}{3}

\defresult{vgg19.ip.snip.0.97}{91.12}
\defresultsd{vgg19.ip.snip.0.97}{0.06}
\defresultn{vgg19.ip.snip.0.97}{3}

\defresult{vgg19.ip.snip.0.99}{90.21}
\defresultsd{vgg19.ip.snip.0.99}{0.16}
\defresultn{vgg19.ip.snip.0.99}{3}

\defresult{vgg19.ip.snip.0.999}{81.47}
\defresultsd{vgg19.ip.snip.0.999}{1.46}
\defresultn{vgg19.ip.snip.0.999}{3}

\defresult{vgg19.ip.wanda.0.95}{91.39}
\defresultsd{vgg19.ip.wanda.0.95}{0.43}
\defresultn{vgg19.ip.wanda.0.95}{3}

\defresult{vgg19.ip.wanda.0.97}{91.13}
\defresultsd{vgg19.ip.wanda.0.97}{0.28}
\defresultn{vgg19.ip.wanda.0.97}{3}

\defresult{vgg19.ip.wanda.0.99}{90.65}
\defresultsd{vgg19.ip.wanda.0.99}{0.42}
\defresultn{vgg19.ip.wanda.0.99}{3}

\defresult{vgg19.ip.wanda.0.999}{84.35}
\defresultsd{vgg19.ip.wanda.0.999}{0.17}
\defresultn{vgg19.ip.wanda.0.999}{3}

\defresult{summary.delta.ncells}{48}
\defresult{summary.delta.npositive}{35}
\defresult{summary.delta.criterion.magnitude}{+2.57}
\defresult{summary.delta.criterion.snip}{+1.29}
\defresult{summary.delta.criterion.wanda}{+6.02}
\defresult{summary.delta.max}{+44.11}
\defresult{summary.delta.mean}{+3.29}
\defresult{summary.delta.min}{-1.58}
\defresult{summary.delta.model.resnet34}{+5.37}
\defresult{summary.delta.model.resnet50}{+7.55}
\defresult{summary.delta.model.vgg11}{-0.35}
\defresult{summary.delta.model.vgg19}{+0.59}
\defresult{summary.delta.sparsity.0.95}{+0.50}
\defresult{summary.delta.sparsity.0.97}{+0.80}
\defresult{summary.delta.sparsity.0.99}{+2.04}
\defresult{summary.delta.sparsity.0.999}{+9.83}

\defresult{resnet34.delta.magnitude.0.95}{+0.83}
\defresult{resnet34.delta.magnitude.0.97}{+1.11}
\defresult{resnet34.delta.magnitude.0.99}{+0.95}
\defresult{resnet34.delta.magnitude.0.999}{+11.80}
\defresult{resnet34.delta.snip.0.95}{+1.62}
\defresult{resnet34.delta.snip.0.97}{+1.39}
\defresult{resnet34.delta.snip.0.99}{+1.45}
\defresult{resnet34.delta.snip.0.999}{+3.20}
\defresult{resnet34.delta.wanda.0.95}{+1.67}
\defresult{resnet34.delta.wanda.0.97}{+2.66}
\defresult{resnet34.delta.wanda.0.99}{+9.00}
\defresult{resnet34.delta.wanda.0.999}{+28.79}
\defresult{resnet50.delta.magnitude.0.95}{+0.39}
\defresult{resnet50.delta.magnitude.0.97}{+0.37}
\defresult{resnet50.delta.magnitude.0.99}{+1.01}
\defresult{resnet50.delta.magnitude.0.999}{+21.82}
\defresult{resnet50.delta.snip.0.95}{+0.77}
\defresult{resnet50.delta.snip.0.97}{+1.35}
\defresult{resnet50.delta.snip.0.99}{+1.55}
\defresult{resnet50.delta.snip.0.999}{+6.00}
\defresult{resnet50.delta.wanda.0.95}{+1.14}
\defresult{resnet50.delta.wanda.0.97}{+1.89}
\defresult{resnet50.delta.wanda.0.99}{+10.26}
\defresult{resnet50.delta.wanda.0.999}{+44.11}
\defresult{vgg11.delta.magnitude.0.95}{-0.46}
\defresult{vgg11.delta.magnitude.0.97}{+0.05}
\defresult{vgg11.delta.magnitude.0.99}{-0.30}
\defresult{vgg11.delta.magnitude.0.999}{-0.42}
\defresult{vgg11.delta.snip.0.95}{-0.11}
\defresult{vgg11.delta.snip.0.97}{+0.08}
\defresult{vgg11.delta.snip.0.99}{-0.06}
\defresult{vgg11.delta.snip.0.999}{-0.32}
\defresult{vgg11.delta.wanda.0.95}{-0.31}
\defresult{vgg11.delta.wanda.0.97}{-0.23}
\defresult{vgg11.delta.wanda.0.99}{-0.50}
\defresult{vgg11.delta.wanda.0.999}{-1.58}
\defresult{vgg19.delta.magnitude.0.95}{+0.23}
\defresult{vgg19.delta.magnitude.0.97}{+0.24}
\defresult{vgg19.delta.magnitude.0.99}{+0.74}
\defresult{vgg19.delta.magnitude.0.999}{+2.72}
\defresult{vgg19.delta.snip.0.95}{+0.28}
\defresult{vgg19.delta.snip.0.97}{+0.28}
\defresult{vgg19.delta.snip.0.99}{+0.42}
\defresult{vgg19.delta.snip.0.999}{+2.74}
\defresult{vgg19.delta.wanda.0.95}{-0.05}
\defresult{vgg19.delta.wanda.0.97}{+0.40}
\defresult{vgg19.delta.wanda.0.99}{-0.03}
\defresult{vgg19.delta.wanda.0.999}{-0.86}

\title{BaCP: Backbone Contrastive Pruning for Preserving Representations in Extremely Sparse Neural Networks}

\author{%
  Mohammad Haroon Khawaja \\
  Lahore University of Management Sciences \\
  Lahore, Pakistan \\
  \texttt{25100225@lums.edu.pk} \\
  \And
  Muhammad Haseeb \\
  Lahore University of Management Sciences \\
  Lahore, Pakistan \\
  \texttt{26100253@lums.edu.pk} \\
  \And
  Mohammad Fatim Shoaib \\
  Lahore University of Management Sciences \\
  Lahore, Pakistan \\
  \texttt{26100004@lums.edu.pk} \\
  \And
  Muhammad Tahir \\
  Lahore University of Management Sciences \\
  Lahore, Pakistan \\
  \texttt{tahir@lums.edu.pk}
}

\begin{document}

\maketitle

\begin{abstract}
Unstructured pruning at extreme sparsity often suffers from representational collapse, causing sharp drops in accuracy. To address this, we study Backbone Contrastive Pruning (BaCP), which regularizes the sparse network's embedding space by aligning it with pretrained, fine-tuned, and historical snapshot models. Building on the contrastive decomposition of the CAP framework~\citep{xu2022dense}, we provide a rigorous matched-budget characterization of this approach across multiple pruning criteria. Evaluated across 90 settings, BaCP improves accuracy substantially in extreme sparsity regimes where standard pruning fails, and is close to baseline where representations remain intact.

\end{abstract}

\section{Introduction}
\label{sec:intro}

While deep neural networks (DNNs) perform well across vision and language tasks, their large number of parameters makes deployment difficult on resource-constrained devices~\citep{wang2025empowering,cheng2024survey}. Network pruning addresses this overhead by removing redundant parameters. Pruning methods are either structured, removing whole components such as channels or attention heads, or unstructured, removing individual weights and allowing much higher sparsity levels. To maximize performance, unstructured pruning is often applied gradually. A sparsity schedule increases sparsity from zero to the target over training, with retraining between pruning steps, so the network can recover as weights are removed~\citep{zhu2017prune,gale2019state}.

While unstructured pruning works well at moderate compression, accuracy degrades sharply at extreme sparsity such as 99.9\%~\citep{li2024pushing}. This has been attributed to representational collapse~\citep{corti2022studying}, since standard methods prune weights based on the network's current state, using metrics like weight magnitudes~\citep{han2015learning}, weight--gradient products~\citep{lee2018snip}, or weight--activation products~\citep{sun2023simple}. As sparsity increases, learned representations can become degraded and unstable, which makes subsequent pruning decisions less reliable.

To address this issue, we study Backbone Contrastive Pruning (BaCP), which regularizes the sparse network's embedding space during pruning. BaCP adds supervised~\citep{khosla2020supervised} and self-supervised~\citep{chen2020simple} contrastive terms to the task loss. These align the sparse model's representations with three frozen references: the pretrained model, the fine-tuned model, and intermediate snapshots. This design follows CAP~\citep{xu2022dense}, which introduced the PrC/FiC/SnC decomposition and its combination with cross-entropy for pruning BERT-family encoders up to 97\% sparsity.

We evaluate BaCP on five convolutional backbones, ResNet-34, ResNet-50, VGG-11, VGG-19, and MobileNetV2, across CIFAR-10 and CIFAR-100, using multiple pruning criteria (Magnitude, SNIP-it, WANDA). All criteria are applied under a single global threshold, keeping the criterion
the only variable between blocks of results. The effect is large under magnitude
and WANDA and near zero under SNIP-it.
We test sparsity up to 99.95\%, beyond the 97\% ceiling reported by CAP. Every BaCP run is matched against an iterative pruning (I.P.) baseline that shares the same initialization, dense checkpoint, criterion, schedule, prunable scope, data splits, pruning epoch budget, and post-pruning fine-tune. The two arms differ in the pruning-phase learning rate and the augmentation stack, and a cross-entropy control holding both of those fixed separates their contribution from the objective's. BaCP's advantage is conditional rather than uniform. It is large where pruning has already degraded the baseline and close to zero where the baseline is intact.

\section{Methodology}
\label{sec:method}

Let $f(\cdot; W)$ be a backbone with weights $W$. Unstructured pruning applies a binary mask $M \in \{0,1\}^{|W|}$, so the active weights are $W \odot M$. For an input $x$ the network produces a feature vector $r = f(x; W \odot M)$, which is routed to two branches. A linear classification head predicts the label. A projection head $g(\cdot)$ maps $r$ to an embedding $z = g(r)/\|g(r)\|_2$ on the unit sphere $\mathbb{S}^{p-1}$. All contrastive terms are computed on these normalized embeddings, where inner products are cosine similarities.

Each model has its own projection head which is a single linear layer of width $p$. The sparse model's head is trainable and is not pruned. The anchor heads are frozen at initialization and are not restored from the anchor checkpoints, so anchor features enter the contrastive space through a fixed random linear map. Projection heads are used only during the pruning phase and are discarded before evaluation, so the final sparse backbone carries no additional inference cost.

\subsection{Anchored Contrastive Alignment}
BaCP regularizes the sparse embedding space by anchoring it to frozen reference models. For each input $x_i$ in a batch of size $B$ we build two augmented views. The sparse model embeds the first view, giving $z_{\mathrm{curr}} \in \mathbb{R}^{B \times p}$, and a frozen anchor $a$ embeds the second, giving $z_{a} \in \mathbb{R}^{B \times p}$. The two are stacked into $Z = [\,z_{\mathrm{curr}} ; z_{a}\,] \in \mathbb{R}^{2B \times p}$ and compared through the full $2B \times 2B$ similarity matrix, so both sparse and anchor embeddings serve as anchors. For an index $i$ and a positive set $P$ we define
\begin{equation}
  \ell(i, P) = -\frac{1}{|P|}\sum_{j \in P}
  \log \frac{\exp\!\big(Z_{i}^{\!\top} Z_{j} / \tau\big)}
            {\sum_{k \neq i} \exp\!\big(Z_{i}^{\!\top} Z_{k} / \tau\big)},
  \label{eq:cap}
\end{equation}
with temperature $\tau = 0.15$. This follows Equation~1 of CAP~\citep{xu2022dense} in form, but over the stacked all-pairs matrix rather than the rectangular sparse-to-anchor block. We evaluate Eq.~\ref{eq:cap} with two positive sets. The instance level set is the other view of the same input~\citep{chen2020simple}. The label level set is $P_y(i) = \{k : y_k = y_i\}$, which covers every sample with the same class~\citep{khosla2020supervised}. The loss for anchor $a$ is
\begin{equation}
  \mathcal{L}_{a} = \frac{1}{2B}\sum_{i=1}^{2B}
  \Big[ \ell\big(i, \mathrm{inst}(i)\big) + \ell\big(i, P_y(i)\big) \Big].
  \label{eq:anchored}
\end{equation}

Anchors are frozen and receive no gradient, so only the sparse model is updated. The mean in Eq.~\ref{eq:anchored} runs over all $2B$ rows, including those of the frozen anchor. Each contrastive term therefore enters Eq.~\ref{eq:total} at roughly half its nominal $\lambda$.

\subsection{Multi-Source Anchors}
\label{sec:anchors}
\begin{figure}[t]
  \centering
    \includegraphics[width=\linewidth]{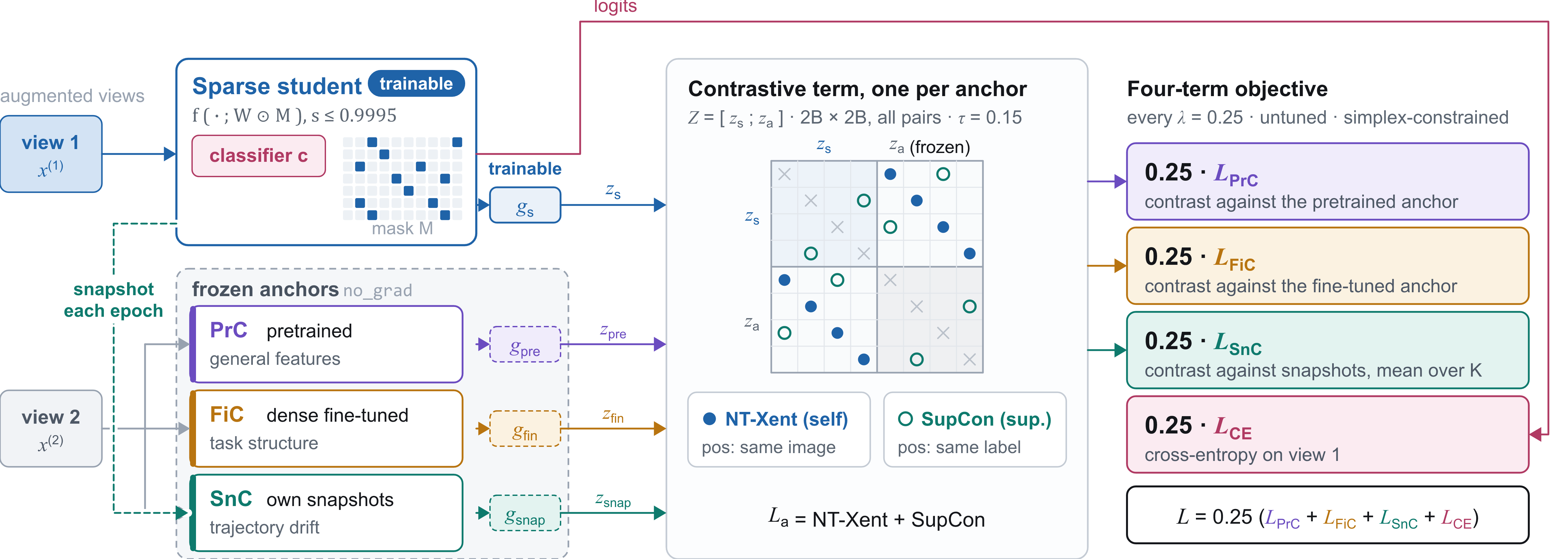}
  \caption{\textbf{BaCP framework.} The sparse model embeds one augmented view and each frozen anchor the other. Every anchor contributes one contrastive term over the stacked similarity matrix.}
  \label{fig:bacp}
\end{figure}

BaCP computes $\mathcal{L}_a$ against three frozen anchors, shown in Figure~\ref{fig:bacp}. The Pretrained Anchor (PrC) is the dense model before task specific fine-tuning and supplies general features. The Fine-tuned Anchor (FiC) is the dense model trained on the target task and supplies class discriminative structure. The Snapshot Anchors (SnC) are $K$ frozen copies of the sparse model saved at earlier pruning stages, and they keep the model consistent with its own recent past across mask updates. The snapshot term is the mean over the saved states, $\mathcal{L}_{\mathrm{SnC}} = \frac{1}{K}\sum_{k=1}^{K} \mathcal{L}_{a}^{(k)}$. We take the mean rather than the sum so that both its scale and $\lambda_{\mathrm{SnC}}$ stay independent of $K$. We retain $K = 2$, since each snapshot costs an additional forward pass per batch.

\subsection{Overall Objective and Pruning Pipeline}
\label{sec:objective}
The sparse model minimizes cross entropy on the classification head together with the three anchored losses:
\begin{equation}
  \mathcal{L} = \lambda_{\mathrm{CE}} \mathcal{L}_{\mathrm{CE}}
  + \lambda_{\mathrm{PrC}} \mathcal{L}_{\mathrm{PrC}}
  + \lambda_{\mathrm{FiC}} \mathcal{L}_{\mathrm{FiC}}
  + \lambda_{\mathrm{SnC}} \mathcal{L}_{\mathrm{SnC}}.
  \label{eq:total}
\end{equation}
Every $\lambda$ is set to $0.25$ and left untuned. When we ablate a component we renormalize the remaining weights to sum to one, so the ablation tests a single change.

\paragraph{Pruning criteria.} BaCP constrains the embedding space and places no requirement on the scoring rule, so any weight criterion can be substituted. We evaluate three standard criteria for the pruning score. Magnitude uses the weight's absolute value, $|w_{ij}|$, and removes the smallest~\citep{han2015learning}. WANDA scores each weight by $|w_{ij}| \lVert X_j \rVert_2$~\citep{sun2023simple} and we threshold it globally along with the other two criteria rather than using its published per-neuron grouping, which does not transfer well to image classifiers. SNIP-it scores each weight by the gradient of the training loss, $|w_{ij}\,\partial \mathcal{L}/\partial w_{ij}|$~\citep{lee2018snip,verdenius2020pruning}. Under BaCP that loss includes the contrastive terms, so the two arms may not prune identical weights. All three criteria threshold at a single global value across the network, so the fraction removed varies by layer and the criteria differ only in which weights the shared budget selects.

\paragraph{Sparsity computation.} We prune the weights of convolutional and linear layers, including the task head, following sparse-training practice that sparsifies every layer rather than exempting the classifier~\citep{evci2020rigging,liu2021granet}. Biases, normalization parameters, and the projection head are never pruned. Sparsity is computed over the pruned layers only and this scope is identical across all compared runs.

\paragraph{Training schedule.} Sparsity follows a cubic schedule~\citep{zhu2017prune},
\begin{equation}
  s_t = s_f\left(1 - \left(1 - \frac{t}{T}\right)^{\!3}\right),
  \label{eq:cubic}
\end{equation}
ramped over the first $80\%$ of steps and frozen at the target $s_f$ for the last $20\%$ as a recovery phase. Masks are recomputed every $\Delta T = 100$ optimizer steps over 50 pruning epochs at batch size 512. Within each step we run the backward pass, update $M$, take the optimizer step, and reapply $M$. Reapplying it after the step prevents momentum and weight decay from reviving pruned weights. Sparsity increases monotonically and pruned weights are zeroed, so masks are nested and no weight is regrown.

\paragraph{Matched baseline.} The iterative pruning baseline is not a separately tuned competitor but the same run with Eq.~\ref{eq:cap} replaced by plain cross entropy. Both runs share the pretrained initialization, the dense checkpoint, the pruning criterion, the cubic schedule, $\Delta T$, the prunable scope, the data splits, the 50 pruning epochs, and the 25-epoch mask-frozen fine-tune. Two elements are not shared. The pruning phase uses SGD at $0.1$ for BaCP and $0.01$ for the baseline, and $0.05$ for BaCP on the VGGs, since the two objectives do not share a gradient scale. BaCP's contrastive terms require two augmented views, so its pruning phase uses a stronger augmentation stack than the baseline's crop and flip. A
cross-entropy control holds both of these at BaCP's settings and zeroes the three contrastive weights,
which separates their effect from the objective's. Under SNIP-it the arms also select different masks, since its score depends on the training loss. Unstated budget differences of this kind are a common failure of pruning comparisons~\citep{blalock2020state}, which is why we state ours.

\section{Results}
\label{sec:results}

Table~\ref{tab:main} reports top-1 test accuracy (\%) on CIFAR-10 under the
protocol of Section~\ref{sec:objective} reported as
$\mathrm{BaCP} = \mathrm{I.P.} + \Delta$. Every entry is the mean of three seeds; per-cell standard
deviations (median 0.29, above 0.3 in 65\% of cells) are in Appendix C.
Across the full grid of 90 cells adding MobileNetV2 and a ResNet-34 CIFAR-100 sweep
(Appendix B) BaCP leads in 53. The margin is concentrated in the cells where
pruning has already damaged the baseline and is close to zero elsewhere.

\begin{table}[t]
\centering
\scriptsize
\setlength{\tabcolsep}{3pt}
\renewcommand{\arraystretch}{0.9}
\caption{Test accuracy (\%) on held-out CIFAR-10, mean of three seeds. BaCP $=$ I.P. $+\ \Delta$; bold marks improvements above $+0.3$. The arms differ in
learning rate and augmentation (Section~\ref{sec:objective}).}
\label{tab:main}
\begin{tabular}{lcc cc cc cc}
\toprule
& \multicolumn{2}{c}{ResNet-34} & \multicolumn{2}{c}{ResNet-50}
& \multicolumn{2}{c}{VGG-11} & \multicolumn{2}{c}{VGG-19} \\
\cmidrule(lr){2-3}\cmidrule(lr){4-5}\cmidrule(lr){6-7}\cmidrule(lr){8-9}
Sparsity & I.P. & $\Delta$ & I.P. & $\Delta$
         & I.P. & $\Delta$ & I.P. & $\Delta$ \\
\midrule
\multicolumn{9}{l}{\textit{Magnitude} \citep{han2015learning}} \\
0.95   & 91.44 & \textbf{+0.36} & 92.82 & +0.10 & 90.04 & \textbf{+0.32} & 91.30 & +0.24 \\
0.97   & 91.51 & +0.15 & 92.65 & $-$0.14 & 90.31 & $-$0.10 & 91.64 & \textbf{+0.40} \\
0.99   & 90.66 & \textbf{+0.38} & 91.59 & +0.01 & 90.10 & $-$0.09 & 91.18 & \textbf{+0.47} \\
0.999  & 76.32 & \textbf{+5.33} & 69.84 & \textbf{+10.76} & 85.31 & $-$0.84 & 84.91 & \textbf{+1.44} \\
0.9995 & 61.09 & \textbf{+14.02} & 38.17 & \textbf{+32.28} & 81.83 & $-$0.88 & 78.70 & \textbf{+2.93} \\
\midrule
\multicolumn{9}{l}{\textit{SNIP-it} \citep{lee2018snip,verdenius2020pruning}} \\
0.95   & 91.26 & \textbf{+0.34} & 92.74 & $-$0.05 & 89.67 & \textbf{+0.34} & 91.44 & $-$0.49 \\
0.97   & 91.02 & +0.28 & 92.26 & +0.03 & 90.07 & $-$0.10 & 91.51 & $-$0.49 \\
0.99   & 90.12 & +0.18 & 91.00 & $-$0.29 & 89.93 & $-$0.36 & 91.31 & $-$1.92 \\
0.999  & 79.92 & $-$0.51 & 78.89 & \textbf{+0.97} & 85.08 & $-$2.09 & 84.49 & $-$3.43 \\
0.9995 & 71.17 & \textbf{+1.85} & 65.44 & \textbf{+5.75} & 81.14 & $-$2.47 & 74.49 & $-$2.94 \\
\midrule
\multicolumn{9}{l}{\textit{WANDA} \citep{sun2023simple}, global threshold} \\
0.95   & 89.65 & \textbf{+1.69} & 92.07 & \textbf{+0.46} & 89.96 & +0.26 & 91.44 & \textbf{+0.55} \\
0.97   & 88.32 & \textbf{+2.88} & 90.92 & \textbf{+1.26} & 90.10 & +0.06 & 91.41 & \textbf{+0.38} \\
0.99   & 81.44 & \textbf{+7.45} & 82.38 & \textbf{+7.74} & 90.13 & $-$0.07 & 91.41 & \textbf{+0.33} \\
0.999  & 38.25 & \textbf{+16.88} & 33.26 & \textbf{+27.65} & 85.39 & $-$0.82 & 86.25 & 0.00 \\
0.9995 & 26.20 & \textbf{+11.73} & 27.62 & \textbf{+4.76} & 81.79 & $-$0.62 & 75.79 & \textbf{+5.03} \\
\bottomrule
\end{tabular}
\end{table}

Excluding 7 collapsed chance-level cells and classifying the 83 live cells by
baseline accuracy makes this explicit. Below 70\% BaCP improves 91\% of cells
with a median margin of $+4.76$ points while above 85\% the median is $+0.01$
and the win rate is 50\%. The damaged-side win rate rises monotonically as the
threshold tightens from 78\% at an 85\% cutoff to 91\% at 70\% and classifying
each cell on seed 1 while measuring the margin on seeds 2 and 3 leaves the effect intact.

A cross-entropy control holds BaCP's augmentation learning rate and schedule
and zeroes the three contrastive weights. In the damaged regime the augmentation and
learning-rate change contributes $+6.62$ points against $+0.77$ for the contrastive
terms. In the intact regime the augmentation costs $-2.08$ and
the objective returns $+1.15$ so the near-zero margin there is two opposing
effects rather than an absence of effect. The same control attributes the MobileNetV2
regression of $-1.77$ to $-5.63$ points at 95\% sparsity (Appendix B) to the
augmentation change.

Appendix C refutes two alternative mechanisms showing projection head capacity does not explain the MobileNetV2 regression and the snapshot schedule leaves the result unchanged.
\section{Conclusion}
\label{sec:conclusion}

BaCP aligns a sparse network's embeddings with frozen pretrained, fine-tuned,
and snapshot references during pruning. Across 90 matched-budget settings it
leads in 53, with the margin concentrated where the baseline is already
damaged: below 70\% baseline accuracy it improves 91\% of cells by a median of
4.76 points, and above 85\% it is indistinguishable from the baseline. A
cross-entropy control attributes most of that gain to the augmentation
and learning-rate change rather than to the objective.

\bibliographystyle{plainnat}
\bibliography{references,methodology_references}

\newpage
\appendix
\raggedbottom
\section{Appendix contents}
\label{app:contents}

\begin{itemize}
  \setlength{\itemsep}{1pt}
  \setlength{\parskip}{0pt}
  \setlength{\parsep}{0pt}
  \setlength{\topsep}{2pt}
  \setlength{\partopsep}{0pt}
  \item[\ref{app:algorithm}] \textbf{Algorithm} --- the procedure in full,
        with the three arms written as settings of one loss line.
  \item[\ref{app:tables}] \textbf{Full result tables} --- absolute top-1 for
        both arms at every one of the 90 grid cells,
        with per-cell seed spread, and the surviving-parameter count that the
        damage account turns on.
  \item[\ref{app:ablations}] \textbf{Complete ablation tables} --- the objective
        (CE-only) decomposition on every cell where it was run, the
        projection-head arms, the snapshot-schedule arms, and the VGG
        learning-rate comparison.
  \item[\ref{app:negative}] \textbf{Negative and refuted results} --- collected
        in one place: the two mechanisms we tested and disproved, the regime
        where the method does nothing, and the cells excluded as collapsed.
  \item[\ref{app:protocol}] \textbf{Protocol as run} --- every hyperparameter,
        stated per family, and what is shared between arms.
  \item[\ref{app:repro}] \textbf{Reproducibility} --- seeds, selection
        signals, and the provenance rules used to admit a run.
  \item[\ref{app:compute}] \textbf{Compute, assets and broader impact} --- the
        hardware, the licenses of everything we used, and what the method does
        and does not protect.
\end{itemize}

\section{Algorithm}
\label{app:algorithm}

Algorithm~\ref{alg:bacp} states the procedure in full. Stages~1--2 and~4 are the
standard prune--retrain template \citep{han2015learning,zhu2017prune}; stage~3 is
where the arms differ, and the only place the objective enters. Throughout,
$E_\theta(\cdot) = \mathrm{normalise}(g_\phi(f_\theta(\cdot)))$ is the embedding
map of Section~\ref{sec:method}.

\begin{algorithm}[H]
\caption{\textbf{Backbone Contrastive Pruning (BaCP).} The three arms are this
same procedure under different settings of line~\ref{alg:line:loss} and the view
count. \textbf{I.P.} sets $\lambda=(0,0,0,1)$ and draws one weakly-augmented
view, so no anchor is evaluated; \textbf{CE-only} sets $\lambda=(0,0,0,1)$ but
keeps the two-view pipeline; \textbf{BaCP} sets
$\lambda=(\tfrac14,\tfrac14,\tfrac14,\tfrac14)$. Everything else is identical
across the three, which is what makes $\Delta$ interpretable.}
\label{alg:bacp}
\begin{algorithmic}[1]
\Require Backbone $f_\theta$, projection head $g_\phi$, classifier $h_\psi$;
         dataset $\mathcal{D}$; target sparsity $s_f$; pruning epochs $E_p$;
         fine-tune epochs $E_f$; mask period $\Delta T$; snapshot cap $K$;
         temperature $\tau$; weights $\lambda=(\lambda_1,\dots,\lambda_4)$;
         scoring rule $\mathrm{score}(\cdot)$
\Ensure  Sparse weights $\theta$ with $\lVert m \rVert_0 / |m| = 1 - s_f$

\Statex \textbf{Stage 1 --- initialisation}
\State $\theta \gets$ ImageNet-pretrained weights; $m \gets \mathbf{1}$
\State $\theta_{\mathrm{pt}} \gets \theta$
       \Comment{frozen pretrained anchor, never updated}

\Statex \textbf{Stage 2 --- dense reference}
\State $\theta_{\mathrm{ft}} \gets \textsc{TrainDense}(\theta, \mathcal{D})$
       \Comment{frozen task anchor; also the checkpoint both arms start from}
\State $\theta \gets \theta_{\mathrm{ft}}$;\quad
       $\mathcal{S} \gets \emptyset$
       \Comment{$\mathcal{S}$: snapshot anchors}

\Statex \textbf{Stage 3 --- gradual pruning under the objective}
\State $T \gets \lfloor 0.8 \cdot E_p \cdot |\mathcal{D}| / B \rfloor$
       \Comment{ramp ends at 80\% of total steps; hold thereafter}
\For{$e = 0$ \textbf{to} $E_p - 1$}
  \For{each minibatch $(x, y)$, at global step $t$}
    \State $x^{(1)}, x^{(2)} \gets \mathrm{aug}(x), \mathrm{aug}(x)$
           \Comment{two independent SimCLR views}
    \State $z \gets E_\theta(x^{(1)})$;\quad
           $\hat y \gets h_\psi(f_\theta(x^{(1)}))$
    \State \textbf{with} no\_grad: \;
           $z_{\mathrm{pt}}, z_{\mathrm{ft}} \gets
            E_{\theta_{\mathrm{pt}}}(x^{(2)}),\; E_{\theta_{\mathrm{ft}}}(x^{(2)})$
    \State $\mathcal{L}_{\mathrm{PrC}} \gets \mathcal{C}(z, z_{\mathrm{pt}}, y)$;\quad
           $\mathcal{L}_{\mathrm{FiC}} \gets \mathcal{C}(z, z_{\mathrm{ft}}, y)$
    \State $\mathcal{L}_{\mathrm{SnC}} \gets
            \begin{cases}
              0 & \mathcal{S} = \emptyset \\[2pt]
              \dfrac{1}{|\mathcal{S}|}\displaystyle\sum_{\theta_k \in \mathcal{S}}
              \mathcal{C}\!\left(z, E_{\theta_k}(x^{(2)}), y\right) & \text{otherwise}
            \end{cases}$
           \Comment{$\mathcal{S}$ is empty for all of epoch 0; averaged, so the
                    scale is independent of $K$}
    \State $\mathcal{L}_{\mathrm{CE}} \gets \mathrm{CE}(\hat y, y)$
    \State $\mathcal{L} \gets \lambda_1\mathcal{L}_{\mathrm{PrC}}
            + \lambda_2\mathcal{L}_{\mathrm{FiC}}
            + \lambda_3\mathcal{L}_{\mathrm{SnC}}
            + \lambda_4\mathcal{L}_{\mathrm{CE}}$
           \label{alg:line:loss}
    \State $\nabla \gets \partial\mathcal{L}/\partial(\theta,\phi,\psi)$;\quad
           clip $\lVert\nabla\rVert_2$ to $10.0$
    \If{$t \bmod \Delta T = 0$ \textbf{and} $t \leq T$}
           \Comment{mask update, on a \emph{step} clock}
      \State $s_t \gets s_f\!\left(1 - \left(1 - t/T\right)^{3}\right)$
      \State $\rho \gets$ the $s_t$-quantile of
             $\{\mathrm{score}(w) : w \in \text{prunable}\}$
             \Comment{one \emph{global} threshold}
      \State $m \gets \mathbf{1}\!\left[\mathrm{score}(\theta) > \rho\right]$
    \EndIf
    \State $(\theta,\phi,\psi) \gets \mathrm{OptStep}(\nabla)$
    \State $\theta \gets \theta \odot m$
           \label{alg:line:mask}
           \Comment{re-imposed \emph{after} the step, or pruned weights drift}
  \EndFor
  \If{$|\mathcal{S}| < K$}
    \State $\mathcal{S} \gets \mathcal{S} \cup \{\mathrm{copy}(\theta)\}$
           \Comment{snapshot anchor, frozen on CPU}
  \EndIf
\EndFor

\Statex \textbf{Stage 4 --- mask-frozen fine-tune}
\State freeze $m$;\quad $g_\phi$ falls out of the graph
       \Comment{stage 4 reads logits only, so the head takes no gradient
                and is never evaluated; it is also never pruned}
\For{$e = 1$ \textbf{to} $E_f$}
  \For{each minibatch $(x,y)$}
    \State $\theta \gets \mathrm{AdamWStep}\!\left(\mathrm{CE}(h_\psi(f_\theta(x)), y)\right)$;\quad
           $\theta \gets \theta \odot m$
           \Comment{no mask updates in this stage}
  \EndFor
\EndFor
\State \Return $\theta \odot m$
\end{algorithmic}
\end{algorithm}

\paragraph{The contrastive functional.}
$\mathcal{C}(z,\hat z,y)$ is CAP's Eq.~1 \citep{xu2022dense} evaluated over the
stacked $2B \times 2B$ similarity matrix on $\mathrm{cat}[z,\hat z]$, as the sum
of a self-supervised instantiation (NT-Xent, $P(i)=\{\hat z_i\}$) and a
supervised one (SupCon, $P(i)=\{\hat z_j : y_j = y_i\}$)
\citep{chen2020simple,khosla2020supervised}:
\begin{equation}
\label{eq:alg-functional}
\mathcal{C}(z,\hat z,y) \;=\; \sum_{i}\; -\frac{1}{|P(i)|}
\sum_{j \in P(i)} \log
\frac{\exp\!\left(z_i^{\top}\hat z_j / \tau\right)}
     {\sum_{k}\exp\!\left(z_i^{\top}\hat z_k / \tau\right)} .
\end{equation}
All three anchors are evaluated under \texttt{no\_grad}, so every term is a
distillation rather than a joint-embedding objective: gradients reach the
student only.

\paragraph{Two details that are easy to state wrongly.}
The mask is re-imposed \emph{after} the optimiser step
(line~\ref{alg:line:mask}), not before: momentum and weight decay would
otherwise walk a pruned weight off zero between mask updates, and the achieved
sparsity would drift below target. And the mask update is gated on the
\emph{global step} index, not the epoch index --- the cubic schedule's endpoint
$T$ is derived from total steps, so driving it on an epoch clock silently
mis-times the ramp. Both are properties of the implementation as run.

\paragraph{Cost.}
Stage~3 runs one student forward-backward plus $2 + |\mathcal{S}|$ frozen
teacher forwards per step. With $K=2$ that is four forwards against I.P.'s one,
which is the price of the objective and is not counted in the epoch budget:
both arms receive $E_p = 50$ pruning epochs and $E_f = 25$ fine-tune epochs, so
$\Delta$ compares equal epochs, not equal wall-clock.

\section{Full result tables}
\label{app:tables}

Table~\ref{tab:app-full} gives absolute top-1 accuracy for both arms at every
grid cell, as mean $\pm$ sample standard deviation over three seeds. The final
column is the number of prunable weights surviving at that sparsity, which is
the quantity that makes the same sparsity ratio mean different things on
different architectures. Greyed $\Delta$ values are the collapsed cells, where
both arms sit at or near chance and the margin is not interpretable; they are
excluded from every aggregate in the main text. Table~\ref{tab:c100} gives the
CIFAR-100 margins in the same form as Table~\ref{tab:main}; the MobileNetV2
rows of the CIFAR-10 grid are in Table~\ref{tab:app-full} with the rest.

\begin{table}[!ht]
\centering\small
\caption{\textbf{ResNet-34 on CIFAR-100}, same protocol. The pattern of Table~\ref{tab:main} reproduces on a second dataset and label space: margins near zero while the baseline is intact, large once it is not.}
\label{tab:c100}
\begin{tabular}{llrrrrr}
\toprule
Model & Criterion & \multicolumn{5}{c}{Sparsity} \\
\cmidrule(lr){3-7}
 &  & 0.95 & 0.97 & 0.99 & 0.999 & 0.9995 \\
\midrule
\textbf{ResNet-34} & magnitude & $+1.82$ & $+1.31$ & $-0.33$ & $+14.08$ & $+11.73$ \\
 & SNIP-it & $+1.87$ & $+1.62$ & $+0.55$ & $-1.40$ & $+0.07$ \\
 & WANDA & $+3.84$ & $+7.70$ & $+18.85$ & $+3.25$ & $+0.63$ \\
\bottomrule
\end{tabular}
\end{table}

\begin{longtable}{llllrrrr}
\caption{\textbf{Every grid cell, both arms.} I.P.\ and BaCP under a matched
50+25-epoch budget, three seeds each.}
\label{tab:app-full} \\
\toprule
Model & Data & Criterion & $s$ & I.P. & BaCP & $\Delta$ & surviving \\
\midrule
\endfirsthead
\toprule
Model & Data & Criterion & $s$ & I.P. & BaCP & $\Delta$ & surviving \\
\midrule
\endhead
\midrule \multicolumn{8}{r}{\emph{continued on next page}} \\
\endfoot
\bottomrule
\endlastfoot
ResNet-34 & C-10 & magnitude & 0.95 & 91.44 {\scriptsize$\pm$0.57} & 91.80 {\scriptsize$\pm$0.21} & $+0.36$ & \num{1063254} \\
ResNet-34 & C-10 & magnitude & 0.97 & 91.51 {\scriptsize$\pm$0.68} & 91.66 {\scriptsize$\pm$0.23} & $+0.15$ & \num{637952} \\
ResNet-34 & C-10 & magnitude & 0.99 & 90.66 {\scriptsize$\pm$0.42} & 91.04 {\scriptsize$\pm$0.53} & $+0.38$ & \num{212650} \\
ResNet-34 & C-10 & magnitude & 0.999 & 76.32 {\scriptsize$\pm$0.67} & 81.65 {\scriptsize$\pm$0.22} & $+5.33$ & \num{21265} \\
ResNet-34 & C-10 & magnitude & 0.9995 & 61.09 {\scriptsize$\pm$1.06} & 75.11 {\scriptsize$\pm$0.14} & $+14.02$ & \num{10632} \\
ResNet-34 & C-10 & SNIP-it & 0.95 & 91.26 {\scriptsize$\pm$0.65} & 91.60 {\scriptsize$\pm$0.14} & $+0.34$ & \num{1063254} \\
ResNet-34 & C-10 & SNIP-it & 0.97 & 91.02 {\scriptsize$\pm$0.45} & 91.30 {\scriptsize$\pm$0.38} & $+0.28$ & \num{637952} \\
ResNet-34 & C-10 & SNIP-it & 0.99 & 90.12 {\scriptsize$\pm$0.52} & 90.30 {\scriptsize$\pm$0.29} & $+0.18$ & \num{212650} \\
ResNet-34 & C-10 & SNIP-it & 0.999 & 79.92 {\scriptsize$\pm$1.12} & 79.41 {\scriptsize$\pm$0.17} & $-0.51$ & \num{21265} \\
ResNet-34 & C-10 & SNIP-it & 0.9995 & 71.17 {\scriptsize$\pm$0.38} & 73.02 {\scriptsize$\pm$0.56} & $+1.85$ & \num{10632} \\
ResNet-34 & C-10 & WANDA & 0.95 & 89.65 {\scriptsize$\pm$0.66} & 91.34 {\scriptsize$\pm$0.04} & $+1.69$ & \num{1063254} \\
ResNet-34 & C-10 & WANDA & 0.97 & 88.32 {\scriptsize$\pm$0.36} & 91.20 {\scriptsize$\pm$0.23} & $+2.88$ & \num{637952} \\
ResNet-34 & C-10 & WANDA & 0.99 & 81.44 {\scriptsize$\pm$0.52} & 88.89 {\scriptsize$\pm$0.04} & $+7.45$ & \num{212650} \\
ResNet-34 & C-10 & WANDA & 0.999 & 38.25 {\scriptsize$\pm$1.17} & 55.13 {\scriptsize$\pm$0.52} & $+16.88$ & \num{21265} \\
ResNet-34 & C-10 & WANDA & 0.9995 & 26.20 {\scriptsize$\pm$0.34} & 37.93 {\scriptsize$\pm$0.85} & $+11.73$ & \num{10632} \\
ResNet-50 & C-10 & magnitude & 0.95 & 92.82 {\scriptsize$\pm$0.42} & 92.92 {\scriptsize$\pm$0.32} & $+0.10$ & \num{1173385} \\
ResNet-50 & C-10 & magnitude & 0.97 & 92.65 {\scriptsize$\pm$0.31} & 92.51 {\scriptsize$\pm$0.57} & $-0.14$ & \num{704031} \\
ResNet-50 & C-10 & magnitude & 0.99 & 91.59 {\scriptsize$\pm$0.27} & 91.60 {\scriptsize$\pm$0.14} & $+0.01$ & \num{234677} \\
ResNet-50 & C-10 & magnitude & 0.999 & 69.84 {\scriptsize$\pm$0.50} & 80.60 {\scriptsize$\pm$0.46} & $+10.76$ & \num{23467} \\
ResNet-50 & C-10 & magnitude & 0.9995 & 38.17 {\scriptsize$\pm$4.76} & 70.45 {\scriptsize$\pm$0.90} & $+32.28$ & \num{11733} \\
ResNet-50 & C-10 & SNIP-it & 0.95 & 92.74 {\scriptsize$\pm$0.44} & 92.69 {\scriptsize$\pm$0.16} & $-0.05$ & \num{1173385} \\
ResNet-50 & C-10 & SNIP-it & 0.97 & 92.26 {\scriptsize$\pm$0.36} & 92.29 {\scriptsize$\pm$0.21} & $+0.03$ & \num{704031} \\
ResNet-50 & C-10 & SNIP-it & 0.99 & 91.00 {\scriptsize$\pm$0.16} & 90.71 {\scriptsize$\pm$0.25} & $-0.29$ & \num{234677} \\
ResNet-50 & C-10 & SNIP-it & 0.999 & 78.89 {\scriptsize$\pm$1.57} & 79.86 {\scriptsize$\pm$0.28} & $+0.97$ & \num{23467} \\
ResNet-50 & C-10 & SNIP-it & 0.9995 & 65.44 {\scriptsize$\pm$1.30} & 71.19 {\scriptsize$\pm$0.39} & $+5.75$ & \num{11733} \\
ResNet-50 & C-10 & WANDA & 0.95 & 92.07 {\scriptsize$\pm$0.48} & 92.53 {\scriptsize$\pm$0.48} & $+0.46$ & \num{1173385} \\
ResNet-50 & C-10 & WANDA & 0.97 & 90.92 {\scriptsize$\pm$0.27} & 92.18 {\scriptsize$\pm$0.35} & $+1.26$ & \num{704031} \\
ResNet-50 & C-10 & WANDA & 0.99 & 82.38 {\scriptsize$\pm$1.88} & 90.12 {\scriptsize$\pm$0.29} & $+7.74$ & \num{234677} \\
ResNet-50 & C-10 & WANDA & 0.999 & 33.26 {\scriptsize$\pm$2.31} & 60.91 {\scriptsize$\pm$1.32} & $+27.65$ & \num{23467} \\
ResNet-50 & C-10 & WANDA & 0.9995 & 27.62 {\scriptsize$\pm$1.69} & 32.38 {\scriptsize$\pm$6.44} & $+4.76$ & \num{11733} \\
VGG-11 & C-10 & magnitude & 0.95 & 90.04 {\scriptsize$\pm$0.15} & 90.36 {\scriptsize$\pm$0.02} & $+0.32$ & \num{6439817} \\
VGG-11 & C-10 & magnitude & 0.97 & 90.31 {\scriptsize$\pm$0.11} & 90.21 {\scriptsize$\pm$0.32} & $-0.10$ & \num{3863890} \\
VGG-11 & C-10 & magnitude & 0.99 & 90.10 {\scriptsize$\pm$0.19} & 90.01 {\scriptsize$\pm$0.21} & $-0.09$ & \num{1287963} \\
VGG-11 & C-10 & magnitude & 0.999 & 85.31 {\scriptsize$\pm$0.12} & 84.47 {\scriptsize$\pm$0.26} & $-0.84$ & \num{128796} \\
VGG-11 & C-10 & magnitude & 0.9995 & 81.83 {\scriptsize$\pm$0.46} & 80.95 {\scriptsize$\pm$0.27} & $-0.88$ & \num{64398} \\
VGG-11 & C-10 & SNIP-it & 0.95 & 89.67 {\scriptsize$\pm$0.16} & 90.01 {\scriptsize$\pm$0.18} & $+0.34$ & \num{6439817} \\
VGG-11 & C-10 & SNIP-it & 0.97 & 90.07 {\scriptsize$\pm$0.20} & 89.97 {\scriptsize$\pm$0.20} & $-0.10$ & \num{3863890} \\
VGG-11 & C-10 & SNIP-it & 0.99 & 89.93 {\scriptsize$\pm$0.29} & 89.57 {\scriptsize$\pm$0.28} & $-0.36$ & \num{1287963} \\
VGG-11 & C-10 & SNIP-it & 0.999 & 85.08 {\scriptsize$\pm$0.03} & 82.99 {\scriptsize$\pm$0.18} & $-2.09$ & \num{128796} \\
VGG-11 & C-10 & SNIP-it & 0.9995 & 81.14 {\scriptsize$\pm$0.56} & 78.67 {\scriptsize$\pm$0.48} & $-2.47$ & \num{64398} \\
VGG-11 & C-10 & WANDA & 0.95 & 89.96 {\scriptsize$\pm$0.06} & 90.22 {\scriptsize$\pm$0.22} & $+0.26$ & \num{6439817} \\
VGG-11 & C-10 & WANDA & 0.97 & 90.10 {\scriptsize$\pm$0.25} & 90.16 {\scriptsize$\pm$0.40} & $+0.06$ & \num{3863890} \\
VGG-11 & C-10 & WANDA & 0.99 & 90.13 {\scriptsize$\pm$0.08} & 90.06 {\scriptsize$\pm$0.17} & $-0.07$ & \num{1287963} \\
VGG-11 & C-10 & WANDA & 0.999 & 85.39 {\scriptsize$\pm$0.13} & 84.57 {\scriptsize$\pm$0.10} & $-0.82$ & \num{128796} \\
VGG-11 & C-10 & WANDA & 0.9995 & 81.79 {\scriptsize$\pm$0.86} & 81.17 {\scriptsize$\pm$0.66} & $-0.62$ & \num{64398} \\
VGG-19 & C-10 & magnitude & 0.95 & 91.30 {\scriptsize$\pm$0.43} & 91.54 {\scriptsize$\pm$0.42} & $+0.24$ & \num{6979875} \\
VGG-19 & C-10 & magnitude & 0.97 & 91.64 {\scriptsize$\pm$0.20} & 92.04 {\scriptsize$\pm$0.29} & $+0.40$ & \num{4187925} \\
VGG-19 & C-10 & magnitude & 0.99 & 91.18 {\scriptsize$\pm$0.10} & 91.65 {\scriptsize$\pm$0.21} & $+0.47$ & \num{1395975} \\
VGG-19 & C-10 & magnitude & 0.999 & 84.91 {\scriptsize$\pm$0.59} & 86.35 {\scriptsize$\pm$0.69} & $+1.44$ & \num{139597} \\
VGG-19 & C-10 & magnitude & 0.9995 & 78.70 {\scriptsize$\pm$1.16} & 81.63 {\scriptsize$\pm$0.71} & $+2.93$ & \num{69798} \\
VGG-19 & C-10 & SNIP-it & 0.95 & 91.44 {\scriptsize$\pm$0.33} & 90.95 {\scriptsize$\pm$0.81} & $-0.49$ & \num{6979875} \\
VGG-19 & C-10 & SNIP-it & 0.97 & 91.51 {\scriptsize$\pm$0.29} & 91.02 {\scriptsize$\pm$0.76} & $-0.49$ & \num{4187925} \\
VGG-19 & C-10 & SNIP-it & 0.99 & 91.31 {\scriptsize$\pm$0.26} & 89.39 {\scriptsize$\pm$3.31} & $-1.92$ & \num{1395975} \\
VGG-19 & C-10 & SNIP-it & 0.999 & 84.49 {\scriptsize$\pm$0.49} & 81.06 {\scriptsize$\pm$4.09} & $-3.43$ & \num{139597} \\
VGG-19 & C-10 & SNIP-it & 0.9995 & 74.49 {\scriptsize$\pm$4.01} & 71.55 {\scriptsize$\pm$1.07} & $-2.94$ & \num{69798} \\
VGG-19 & C-10 & WANDA & 0.95 & 91.44 {\scriptsize$\pm$0.16} & 91.99 {\scriptsize$\pm$0.16} & $+0.55$ & \num{6979875} \\
VGG-19 & C-10 & WANDA & 0.97 & 91.41 {\scriptsize$\pm$0.13} & 91.79 {\scriptsize$\pm$0.27} & $+0.38$ & \num{4187925} \\
VGG-19 & C-10 & WANDA & 0.99 & 91.41 {\scriptsize$\pm$0.22} & 91.74 {\scriptsize$\pm$0.02} & $+0.33$ & \num{1395975} \\
VGG-19 & C-10 & WANDA & 0.999 & 86.25 {\scriptsize$\pm$0.18} & 86.25 {\scriptsize$\pm$1.07} & $0.00$ & \num{139597} \\
VGG-19 & C-10 & WANDA & 0.9995 & 75.79 {\scriptsize$\pm$6.59} & 80.82 {\scriptsize$\pm$0.42} & $+5.03$ & \num{69798} \\
MobileNetV2 & C-10 & magnitude & 0.95 & 89.72 {\scriptsize$\pm$0.23} & 87.95 {\scriptsize$\pm$0.40} & $-1.77$ & \num{110128} \\
MobileNetV2 & C-10 & magnitude & 0.97 & 88.87 {\scriptsize$\pm$0.42} & 86.27 {\scriptsize$\pm$0.08} & $-2.60$ & \num{66076} \\
MobileNetV2 & C-10 & magnitude & 0.99 & 85.73 {\scriptsize$\pm$0.22} & 80.81 {\scriptsize$\pm$0.34} & $-4.92$ & \num{22025} \\
MobileNetV2 & C-10 & magnitude & 0.999 & 10.00 {\scriptsize$\pm$0.00} & 10.00 {\scriptsize$\pm$0.00} & \textcolor{gray}{$0.00$} & \num{2202} \\
MobileNetV2 & C-10 & magnitude & 0.9995 & 10.00 {\scriptsize$\pm$0.00} & 10.00 {\scriptsize$\pm$0.00} & \textcolor{gray}{$0.00$} & \num{1101} \\
MobileNetV2 & C-10 & SNIP-it & 0.95 & 90.33 {\scriptsize$\pm$0.17} & 87.80 {\scriptsize$\pm$0.11} & $-2.53$ & \num{110128} \\
MobileNetV2 & C-10 & SNIP-it & 0.97 & 89.28 {\scriptsize$\pm$0.10} & 86.06 {\scriptsize$\pm$0.17} & $-3.22$ & \num{66076} \\
MobileNetV2 & C-10 & SNIP-it & 0.99 & 85.41 {\scriptsize$\pm$0.19} & 80.09 {\scriptsize$\pm$0.29} & $-5.32$ & \num{22025} \\
MobileNetV2 & C-10 & SNIP-it & 0.999 & 10.00 {\scriptsize$\pm$0.00} & 24.05 {\scriptsize$\pm$3.81} & \textcolor{gray}{$+14.05$} & \num{2202} \\
MobileNetV2 & C-10 & SNIP-it & 0.9995 & 10.00 {\scriptsize$\pm$0.00} & 10.00 {\scriptsize$\pm$0.00} & \textcolor{gray}{$0.00$} & \num{1101} \\
MobileNetV2 & C-10 & WANDA & 0.95 & 89.24 {\scriptsize$\pm$0.27} & 83.61 {\scriptsize$\pm$1.29} & $-5.63$ & \num{110128} \\
MobileNetV2 & C-10 & WANDA & 0.97 & 87.64 {\scriptsize$\pm$0.04} & 69.08 {\scriptsize$\pm$19.75} & $-18.56$ & \num{66076} \\
MobileNetV2 & C-10 & WANDA & 0.99 & 10.00 {\scriptsize$\pm$0.00} & 10.00 {\scriptsize$\pm$0.00} & \textcolor{gray}{$0.00$} & \num{22025} \\
MobileNetV2 & C-10 & WANDA & 0.999 & 10.00 {\scriptsize$\pm$0.00} & 10.00 {\scriptsize$\pm$0.00} & \textcolor{gray}{$0.00$} & \num{2202} \\
MobileNetV2 & C-10 & WANDA & 0.9995 & 10.00 {\scriptsize$\pm$0.00} & 10.00 {\scriptsize$\pm$0.00} & \textcolor{gray}{$0.00$} & \num{1101} \\
ResNet-34 & C-100 & magnitude & 0.95 & 69.91 {\scriptsize$\pm$1.14} & 71.73 {\scriptsize$\pm$0.13} & $+1.82$ & \num{1063254} \\
ResNet-34 & C-100 & magnitude & 0.97 & 69.74 {\scriptsize$\pm$1.65} & 71.05 {\scriptsize$\pm$0.28} & $+1.31$ & \num{637952} \\
ResNet-34 & C-100 & magnitude & 0.99 & 67.77 {\scriptsize$\pm$1.26} & 67.44 {\scriptsize$\pm$0.75} & $-0.33$ & \num{212650} \\
ResNet-34 & C-100 & magnitude & 0.999 & 29.35 {\scriptsize$\pm$1.31} & 43.43 {\scriptsize$\pm$0.25} & $+14.08$ & \num{21265} \\
ResNet-34 & C-100 & magnitude & 0.9995 & 13.23 {\scriptsize$\pm$0.44} & 24.96 {\scriptsize$\pm$1.69} & $+11.73$ & \num{10632} \\
ResNet-34 & C-100 & SNIP-it & 0.95 & 69.06 {\scriptsize$\pm$1.10} & 70.93 {\scriptsize$\pm$0.49} & $+1.87$ & \num{1063254} \\
ResNet-34 & C-100 & SNIP-it & 0.97 & 68.47 {\scriptsize$\pm$1.39} & 70.09 {\scriptsize$\pm$0.47} & $+1.62$ & \num{637952} \\
ResNet-34 & C-100 & SNIP-it & 0.99 & 64.84 {\scriptsize$\pm$0.37} & 65.39 {\scriptsize$\pm$0.39} & $+0.55$ & \num{212650} \\
ResNet-34 & C-100 & SNIP-it & 0.999 & 36.83 {\scriptsize$\pm$0.51} & 35.43 {\scriptsize$\pm$0.62} & $-1.40$ & \num{21265} \\
ResNet-34 & C-100 & SNIP-it & 0.9995 & 20.91 {\scriptsize$\pm$0.99} & 20.98 {\scriptsize$\pm$0.39} & $+0.07$ & \num{10632} \\
ResNet-34 & C-100 & WANDA & 0.95 & 66.43 {\scriptsize$\pm$0.65} & 70.27 {\scriptsize$\pm$0.50} & $+3.84$ & \num{1063254} \\
ResNet-34 & C-100 & WANDA & 0.97 & 61.52 {\scriptsize$\pm$0.73} & 69.22 {\scriptsize$\pm$0.95} & $+7.70$ & \num{637952} \\
ResNet-34 & C-100 & WANDA & 0.99 & 41.57 {\scriptsize$\pm$0.80} & 60.42 {\scriptsize$\pm$0.18} & $+18.85$ & \num{212650} \\
ResNet-34 & C-100 & WANDA & 0.999 & 3.40 {\scriptsize$\pm$1.12} & 6.65 {\scriptsize$\pm$0.80} & $+3.25$ & \num{21265} \\
ResNet-34 & C-100 & WANDA & 0.9995 & 2.09 {\scriptsize$\pm$0.35} & 2.72 {\scriptsize$\pm$0.32} & $+0.63$ & \num{10632} \\
\end{longtable}

\section{Complete ablation tables}
\label{app:ablations}

\subsection{Objective ablation, every cell}
The cross-entropy control of Section~\ref{sec:results}, written CE-only in the
tables below, holds every BaCP setting and weights the three contrastive terms
to zero, so the reported margin decomposes exactly into
$(\mathrm{CE\text{-}only}-\mathrm{I.P.})$, the augmentation and learning-rate
change, plus $(\mathrm{BaCP}-\mathrm{CE\text{-}only})$, the objective.

\begin{longtable}{llrrrrrrl}
\caption{\textbf{Objective ablation}, all cells, three seeds per arm.}
\label{tab:app-obj} \\
\toprule
Model & $s$ & Criterion & I.P. & CE-only & BaCP & aug$+$LR & contrastive & regime \\
\midrule
\endfirsthead
\toprule
Model & $s$ & Criterion & I.P. & CE-only & BaCP & aug$+$LR & contrastive & regime \\
\midrule
\endhead
\bottomrule
\endlastfoot
MobileNetV2 & 0.99 & magnitude & 85.73 {\scriptsize$\pm$0.22} & 79.55 {\scriptsize$\pm$1.44} & 80.81 {\scriptsize$\pm$0.34} & $-6.18$ & $+1.26$ & intact \\
ResNet-34 & 0.95 & magnitude & 91.44 {\scriptsize$\pm$0.57} & 90.45 {\scriptsize$\pm$0.42} & 91.80 {\scriptsize$\pm$0.21} & $-0.99$ & $+1.35$ & intact \\
ResNet-34 & 0.999 & magnitude & 76.32 {\scriptsize$\pm$0.67} & 80.09 {\scriptsize$\pm$1.04} & 81.65 {\scriptsize$\pm$0.22} & $+3.77$ & $+1.56$ & damaged \\
ResNet-34 & 0.999 & SNIP-it & 79.92 {\scriptsize$\pm$1.12} & 77.67 {\scriptsize$\pm$0.77} & 79.41 {\scriptsize$\pm$0.17} & $-2.25$ & $+1.74$ & damaged \\
ResNet-34 & 0.999 & WANDA & 38.25 {\scriptsize$\pm$1.17} & 56.25 {\scriptsize$\pm$3.01} & 55.13 {\scriptsize$\pm$0.52} & $+18.00$ & $-1.12$ & damaged \\
ResNet-50 & 0.95 & magnitude & 92.82 {\scriptsize$\pm$0.42} & 91.55 {\scriptsize$\pm$0.00} & 92.92 {\scriptsize$\pm$0.32} & $-1.27$ & $+1.37$ & intact \\
ResNet-50 & 0.999 & magnitude & 69.84 {\scriptsize$\pm$0.50} & 77.44 {\scriptsize$\pm$1.65} & 80.60 {\scriptsize$\pm$0.46} & $+7.60$ & $+3.16$ & damaged \\
ResNet-50 & 0.999 & SNIP-it & 78.89 {\scriptsize$\pm$1.57} & 77.56 {\scriptsize$\pm$0.50} & 79.86 {\scriptsize$\pm$0.28} & $-1.33$ & $+2.30$ & damaged \\
ResNet-50 & 0.999 & WANDA & 33.26 {\scriptsize$\pm$2.31} & 60.51 {\scriptsize$\pm$2.23} & 60.91 {\scriptsize$\pm$1.32} & $+27.25$ & $+0.40$ & damaged \\
VGG-11 & 0.95 & magnitude & 90.04 {\scriptsize$\pm$0.15} & 90.09 {\scriptsize$\pm$0.26} & 90.36 {\scriptsize$\pm$0.02} & $+0.05$ & $+0.27$ & intact \\
VGG-11 & 0.999 & magnitude & 85.31 {\scriptsize$\pm$0.12} & 86.31 {\scriptsize$\pm$0.24} & 84.47 {\scriptsize$\pm$0.26} & $+1.00$ & $-1.84$ & intact \\
VGG-11 & 0.999 & SNIP-it & 85.08 {\scriptsize$\pm$0.03} & 83.83 {\scriptsize$\pm$0.90} & 82.99 {\scriptsize$\pm$0.18} & $-1.25$ & $-0.84$ & intact \\
VGG-11 & 0.999 & WANDA & 85.39 {\scriptsize$\pm$0.13} & 86.04 {\scriptsize$\pm$0.15} & 84.57 {\scriptsize$\pm$0.10} & $+0.65$ & $-1.47$ & intact \\
VGG-19 & 0.999 & magnitude & 84.91 {\scriptsize$\pm$0.59} & 88.29 {\scriptsize$\pm$0.26} & 86.35 {\scriptsize$\pm$0.69} & $+3.38$ & $-1.94$ & damaged \\
VGG-19 & 0.95 & SNIP-it & 91.44 {\scriptsize$\pm$0.33} & 80.74 {\scriptsize$\pm$6.64} & 90.95 {\scriptsize$\pm$0.81} & $-10.70$ & $+10.21$ & intact \\
VGG-19 & 0.999 & SNIP-it & 84.49 {\scriptsize$\pm$0.49} & 81.01 {\scriptsize$\pm$1.83} & 81.06 {\scriptsize$\pm$4.09} & $-3.48$ & $+0.05$ & damaged \\
VGG-19 & 0.999 & WANDA & 86.25 {\scriptsize$\pm$0.18} & 86.18 {\scriptsize$\pm$1.18} & 86.25 {\scriptsize$\pm$1.07} & $-0.07$ & $+0.07$ & intact \\
\end{longtable}

\subsection{Projection head}
The head is $\mathrm{Linear}(d,128)$, never pruned and discarded at evaluation.
\texttt{headfrozen} shares one frozen head across student and teachers;
\texttt{headnone} removes the head and contrasts pooled backbone features.

\begin{table}[h]
\centering\small
\begin{tabular}{lllrrr}
\toprule
Arm & Model & $s$ & acc. & seeds & vs.\ control \\
\midrule
\texttt{head128} & MobileNetV2 & 0.95 & 87.95 & 3 & control \\
\texttt{head16} & MobileNetV2 & 0.95 & 87.08 & 1 & $-0.87$ \\
\texttt{head32} & MobileNetV2 & 0.95 & 87.56 & 1 & $-0.39$ \\
\texttt{head64} & MobileNetV2 & 0.95 & 87.74 & 1 & $-0.21$ \\
\texttt{headfrozen} & MobileNetV2 & 0.95 & 87.74 & 1 & $-0.21$ \\
\texttt{headnone} & MobileNetV2 & 0.95 & 87.95 & 1 & $0.00$ \\
\texttt{head128} & MobileNetV2 & 0.99 & 80.81 & 3 & control \\
\texttt{head16} & MobileNetV2 & 0.99 & 80.29 & 1 & $-0.52$ \\
\texttt{head32} & MobileNetV2 & 0.99 & 80.48 & 1 & $-0.33$ \\
\texttt{head64} & MobileNetV2 & 0.99 & 80.75 & 1 & $-0.06$ \\
\texttt{headfrozen} & MobileNetV2 & 0.99 & 79.99 & 1 & $-0.82$ \\
\texttt{head128} & ResNet-34 & 0.999 & 81.65 & 3 & control \\
\texttt{head16} & ResNet-34 & 0.999 & 81.11 & 1 & $-0.54$ \\
\bottomrule
\end{tabular}
\end{table}

\subsection{Snapshot schedule}
\texttt{snapfirst2} is the protocol setting: the cap is filled greedily in the
opening epochs. \texttt{snapeven$K$} spreads $K$ snapshots evenly across the
pruning phase.

\begin{table}[h]
\centering\small
\begin{tabular}{lllrr}
\toprule
Arm & Model & $s$ & acc. & vs.\ control \\
\midrule
\texttt{snapeven2} & ResNet-34 & 0.97 & 92.32 & $+0.28$ \\
\texttt{snapeven3} & ResNet-34 & 0.97 & 92.17 & $+0.13$ \\
\texttt{snapeven5} & ResNet-34 & 0.97 & 92.03 & $-0.01$ \\
\texttt{snapfirst2} & ResNet-34 & 0.97 & 92.04 & $0.00$ \\
\texttt{snapeven2} & ResNet-34 & 0.999 & 81.14 & $-0.26$ \\
\texttt{snapeven3} & ResNet-34 & 0.999 & 81.19 & $-0.21$ \\
\texttt{snapeven5} & ResNet-34 & 0.999 & 81.68 & $+0.28$ \\
\texttt{snapfirst2} & ResNet-34 & 0.999 & 81.40 & $0.00$ \\
\bottomrule
\end{tabular}
\end{table}

\subsection{VGG learning rate}
The only architecture-specific choice in the protocol, measured rather than
asserted. Three seeds per entry.

\begin{table}[h]
\centering\small
\begin{tabular}{llrrrr}
\toprule
Model & $s$ & I.P.\ @\,0.01 & BaCP @\,0.05 & BaCP @\,0.1 & 0.1 vs 0.05 \\
\midrule
VGG-11 & 0.95 & 90.04 {\scriptsize$\pm$0.15} & 90.36 {\scriptsize$\pm$0.02} & 87.29 {\scriptsize$\pm$1.92} & $-3.07$ \\
VGG-11 & 0.999 & 85.31 {\scriptsize$\pm$0.12} & 84.47 {\scriptsize$\pm$0.26} & 83.05 {\scriptsize$\pm$2.00} & $-1.42$ \\
\bottomrule
\end{tabular}
\end{table}

\section{Negative and refuted results}
\label{app:negative}

Collected here rather than distributed through the main text, so that a reader
can see in one place what did not work.

\paragraph{The method does nothing in the intact regime.}
Above an 85\% baseline the margin has a median of approximately zero and a win
rate indistinguishable from chance across 46 cells. This is half the grid. We report it as a scope condition, not as an
inconvenience: it is the same finding as the damaged-regime result, read from
the other side.

\paragraph{The projection-head hypothesis is false.}
The head reaches 744\% of the surviving backbone on MobileNetV2 at 0.99, which
makes it the obvious explanation for that model's regression. Nine comparisons
across two sparsities --- three narrower heads, a frozen head, and no head at
all --- produced no improvement over the default. The capacity ratio is real
and not causal.

\paragraph{The snapshot schedule is irrelevant.}
Three alternative schedules at two sparsities moved the result by at most a few
tenths of a point, in no consistent direction.

\paragraph{Matching VGG's learning rate to the other families is harmful.}
Raising the VGG-11 pruning-phase rate from 0.05 to 0.1 costs accuracy at both
sparsities tested and raises the seed spread by an order of magnitude.

\paragraph{Collapsed cells.}
The following cells have both arms at or near chance. A margin between two
destroyed networks measures nothing, so they are excluded from every aggregate.

\begin{table}[h]
\centering\small
\begin{tabular}{llllrr}
\toprule
Model & Data & Criterion & $s$ & I.P. & BaCP \\
\midrule
MobileNetV2 & C-10 & magnitude & 0.999 & 10.00 & 10.00 \\
MobileNetV2 & C-10 & magnitude & 0.9995 & 10.00 & 10.00 \\
MobileNetV2 & C-10 & SNIP-it & 0.999 & 10.00 & 24.05 \\
MobileNetV2 & C-10 & SNIP-it & 0.9995 & 10.00 & 10.00 \\
MobileNetV2 & C-10 & WANDA & 0.99 & 10.00 & 10.00 \\
MobileNetV2 & C-10 & WANDA & 0.999 & 10.00 & 10.00 \\
MobileNetV2 & C-10 & WANDA & 0.9995 & 10.00 & 10.00 \\
\bottomrule
\end{tabular}
\end{table}

\section{Protocol as run}
\label{app:protocol}

\begin{table}[h]
\centering\small
\begin{tabular}{lll}
\toprule
 & I.P. & BaCP \\
\midrule
Initialisation & \multicolumn{2}{c}{ImageNet-pretrained, identical checkpoint} \\
Dense reference & \multicolumn{2}{c}{identical, per seed} \\
Criterion & \multicolumn{2}{c}{identical (magnitude / SNIP-it / WANDA, global threshold)} \\
Sparsity schedule & \multicolumn{2}{c}{cubic, ramp over 80\% of steps, $\Delta T=100$} \\
Prunable scope & \multicolumn{2}{c}{identical; classifier head pruned, projection head never} \\
Pruning epochs & 50 & 50 \\
Fine-tune & \multicolumn{2}{c}{25 epochs, mask frozen, AdamW $10^{-4}$} \\
Batch size & \multicolumn{2}{c}{512} \\
Data splits & \multicolumn{2}{c}{45\,000 / 5\,000 / 10\,000, identical} \\
Precision & \multicolumn{2}{c}{bf16 autocast} \\
Gradient clipping & \multicolumn{2}{c}{global norm 10.0} \\
\midrule
Objective & cross-entropy & $\sum_i \lambda_i \mathcal{L}_i$, $\lambda=(.25,.25,.25,.25)$ \\
Augmentation & single weak view & two SimCLR views \\
Pruning-phase LR & SGD 0.01 & SGD 0.1 (0.05 on VGG) \\
Temperature & --- & $\tau = 0.15$ \\
Snapshots & --- & $K=2$, greedy \\
\bottomrule
\end{tabular}
\end{table}

The three rows below the rule are the differences. The cross-entropy control of
Section~\ref{sec:results} holds the augmentation and learning-rate rows at
BaCP's values and the objective row at cross-entropy, which is what separates
their contribution from the objective's exactly rather than approximately.

\section{Reproducibility}
\label{app:repro}

Every cell was run at seeds 1, 2 and 3 on both arms; no cell is reported from a
single seed except the head and snapshot ablations, which are labelled as such
and are used only to refute mechanisms, never to set a hyperparameter. Selection
used training loss for the contrastive phase (which has no supervised validation
signal), validation accuracy for both fine-tunes, and the test set for nothing.

A run was admitted to the grid only if its recorded configuration matched the
protocol exactly: 50 pruning epochs, 25 fine-tune epochs, the declared sparsity,
criterion and seed, and a verified load of the dense checkpoint. Runs failing any
of these --- including an earlier generation in which the contrastive student
silently failed to load the dense checkpoint --- were excluded by an automated
check over the recorded fields rather than by hand.

\section{Compute, assets and broader impact}
\label{app:compute}

\paragraph{Compute.}
Every run reported here was executed on a single NVIDIA A100 GPU on a cloud node
with 24 vCPUs (an Azure \texttt{Standard\_NC24ads\_A100\_v4} instance). Runs are
serialised, one training job per GPU, and nothing in this work is multi-GPU or
multi-node. The grid is 540 runs: 90 cells at three seeds on each of the two
arms. Cells are not uniform in cost. A BaCP step forwards two augmented views
through the student and $2 + |\mathcal{S}|$ frozen anchors, which at $K = 2$ is
four forward passes against I.P.'s one, so a BaCP cell is several times more
expensive per epoch than the baseline it is compared against. The matched-budget
claim is a claim about epochs and schedule, which are identical across arms; it
is not a claim that the two arms cost the same GPU-time, and they do not. The
project also consumed compute that is not reported here: an earlier generation
of runs was discarded after the provenance check described above rejected them,
and the protocol probes of Appendix~\ref{app:negative} were run at one seed.

\paragraph{Existing assets.}
CIFAR-10 and CIFAR-100 \citep{krizhevsky2009learning} are used under their
standard research release; no license is attached by the authors, and the
datasets are used as distributed, unmodified, for their intended purpose of
image-classification benchmarking. Backbones are the \texttt{torchvision}
checkpoints for ResNet-34/50, VGG-11/19 and MobileNetV2
\citep{he2016deep,simonyan2014very,sandler2018mobilenetv2}, pretrained on
ImageNet \citep{deng2009imagenet} and used under torchvision's BSD-3-Clause
license. Training uses PyTorch, also BSD-3-Clause, with SGD for the pruning
phase and AdamW \citep{loshchilov2019adamw} for the mask-frozen fine-tune.
The pruning criteria are reimplemented from their published descriptions
\citep{han2015learning,lee2018snip,sun2023simple} rather than taken from the
authors' code. No new dataset or model is released with this paper.

\paragraph{Broader impact.}
This work makes an existing class of sparse networks slightly more accurate at a
fixed parameter budget. The positive case is the usual one for pruning: a
network that keeps its accuracy at 99.9\% sparsity is cheaper to store and to
run, which lowers the energy and hardware cost of deployment and widens the
range of devices a model can run on. We are aware of no path from this work to a
directly harmful application that was not already open. Making inference cheaper
marginally lowers the cost of deploying any classifier, benign or otherwise, but
the models studied here are standard published architectures on a standard
public benchmark, and nothing about the objective is specific to a sensitive
use. One caveat worth stating is that pruning is known to affect classes
unevenly \citep{hooker2019compressed}: a compressed model can lose accuracy on
under-represented groups faster than its aggregate accuracy suggests. We report
aggregate top-1 only and do not measure per-class or per-group degradation, so
nothing here should be read as evidence that BaCP protects tail classes.

\newpage
\section*{NeurIPS Paper Checklist}

The checklist is designed to encourage best practices for responsible machine learning research, addressing issues of reproducibility, transparency, research ethics, and societal impact. Do not remove the checklist: {\bf The papers not including the checklist will be desk rejected.} The checklist should follow the references and follow the (optional) supplemental material.  The checklist does NOT count towards the page
limit. 

Please read the checklist guidelines carefully for information on how to answer these questions. For each question in the checklist:
\begin{itemize}
    \item You should answer \answerYes{}, \answerNo{}, or \answerNA{}.
    \item \answerNA{} means either that the question is Not Applicable for that particular paper or the relevant information is Not Available.
    \item Please provide a short (1--2 sentence) justification right after your answer (even for \answerNA). 
\end{itemize}

{\bf The checklist answers are an integral part of your paper submission.} They are visible to the reviewers, area chairs, senior area chairs, and ethics reviewers. You will also be asked to include it (after eventual revisions) with the final version of your paper, and its final version will be published with the paper.

The reviewers of your paper will be asked to use the checklist as one of the factors in their evaluation. While \answerYes{} is generally preferable to \answerNo{}, it is perfectly acceptable to answer \answerNo{} provided a proper justification is given (e.g., error bars are not reported because it would be too computationally expensive'' or ``we were unable to find the license for the dataset we used''). In general, answering \answerNo{} or \answerNA{} is not grounds for rejection. While the questions are phrased in a binary way, we acknowledge that the true answer is often more nuanced, so please just use your best judgment and write a justification to elaborate. All supporting evidence can appear either in the main paper or the supplemental material, provided in appendix. If you answer \answerYes{} to a question, in the justification please point to the section(s) where related material for the question can be found.

IMPORTANT, please:
\begin{itemize}
    \item {\bf Delete this instruction block, but keep the section heading ``NeurIPS Paper Checklist"},
    \item  {\bf Keep the checklist subsection headings, questions/answers and guidelines below.}
    \item {\bf Do not modify the questions and only use the provided macros for your answers}.
\end{itemize}


\begin{enumerate}

\item {\bf Claims}
    \item[] Question: Do the main claims made in the abstract and introduction accurately reflect the paper's contributions and scope?
    \item[] Answer: \answerYes{} 
    \item[] Justification: The abstract and Section~\ref{sec:intro} both state the claim in its conditional form: the advantage is large where pruning has already damaged the baseline and close to zero where it has not. Section~\ref{sec:results} gives the number the claim rests on, 53 of 90 cells, before the conditional split, and the cross-entropy control there attributes only about a tenth of the damaged-regime margin to the objective itself. No unconditional improvement is claimed anywhere.
    \item[] Guidelines:
    \begin{itemize}
        \item The answer \answerNA{} means that the abstract and introduction do not include the claims made in the paper.
        \item The abstract and/or introduction should clearly state the claims made, including the contributions made in the paper and important assumptions and limitations. A \answerNo{} or \answerNA{} answer to this question will not be perceived well by the reviewers. 
        \item The claims made should match theoretical and experimental results, and reflect how much the results can be expected to generalize to other settings. 
        \item It is fine to include aspirational goals as motivation as long as it is clear that these goals are not attained by the paper. 
    \end{itemize}

\item {\bf Limitations}
    \item[] Question: Does the paper discuss the limitations of the work performed by the authors?
    \item[] Answer: \answerYes{} 
    \item[] Justification: Section~\ref{sec:results} reports that the method does nothing above an 85\% baseline, that VGG-11 never enters the regime where it helps, and that MobileNetV2 regresses; the cross-entropy control shows roughly nine tenths of the damaged-regime margin comes from the augmentation and learning-rate change rather than the objective. Appendix~\ref{app:negative} collects the negative and refuted results in one place, and Appendix~\ref{app:compute} notes that we report aggregate top-1 only and do not measure per-class degradation.
    \item[] Guidelines:
    \begin{itemize}
        \item The answer \answerNA{} means that the paper has no limitation while the answer \answerNo{} means that the paper has limitations, but those are not discussed in the paper. 
        \item The authors are encouraged to create a separate ``Limitations'' section in their paper.
        \item The paper should point out any strong assumptions and how robust the results are to violations of these assumptions (e.g., independence assumptions, noiseless settings, model well-specification, asymptotic approximations only holding locally). The authors should reflect on how these assumptions might be violated in practice and what the implications would be.
        \item The authors should reflect on the scope of the claims made, e.g., if the approach was only tested on a few datasets or with a few runs. In general, empirical results often depend on implicit assumptions, which should be articulated.
        \item The authors should reflect on the factors that influence the performance of the approach. For example, a facial recognition algorithm may perform poorly when image resolution is low or images are taken in low lighting. Or a speech-to-text system might not be used reliably to provide closed captions for online lectures because it fails to handle technical jargon.
        \item The authors should discuss the computational efficiency of the proposed algorithms and how they scale with dataset size.
        \item If applicable, the authors should discuss possible limitations of their approach to address problems of privacy and fairness.
        \item While the authors might fear that complete honesty about limitations might be used by reviewers as grounds for rejection, a worse outcome might be that reviewers discover limitations that aren't acknowledged in the paper. The authors should use their best judgment and recognize that individual actions in favor of transparency play an important role in developing norms that preserve the integrity of the community. Reviewers will be specifically instructed to not penalize honesty concerning limitations.
    \end{itemize}

\item {\bf Theory assumptions and proofs}
    \item[] Question: For each theoretical result, does the paper provide the full set of assumptions and a complete (and correct) proof?
    \item[] Answer: \answerNA{} 
    \item[] Justification: The paper contains no theorems, lemmas or formal claims. The equations state the training objective and the sparsity schedule and are definitions, not results requiring proof.
    \item[] Guidelines:
    \begin{itemize}
        \item The answer \answerNA{} means that the paper does not include theoretical results. 
        \item All the theorems, formulas, and proofs in the paper should be numbered and cross-referenced.
        \item All assumptions should be clearly stated or referenced in the statement of any theorems.
        \item The proofs can either appear in the main paper or the supplemental material, but if they appear in the supplemental material, the authors are encouraged to provide a short proof sketch to provide intuition. 
        \item Inversely, any informal proof provided in the core of the paper should be complemented by formal proofs provided in appendix or supplemental material.
        \item Theorems and Lemmas that the proof relies upon should be properly referenced. 
    \end{itemize}

    \item {\bf Experimental result reproducibility}
    \item[] Question: Does the paper fully disclose all the information needed to reproduce the main experimental results of the paper to the extent that it affects the main claims and/or conclusions of the paper (regardless of whether the code and data are provided or not)?
    \item[] Answer: \answerYes{} 
    \item[] Justification: Section~\ref{sec:objective} gives the objective, the criteria, the prunable scope, the schedule and the matched-baseline protocol. Appendix~\ref{app:algorithm} states the procedure as pseudocode, including the two implementation details that are easy to get wrong: the mask is re-imposed after the optimiser step, and the mask update is gated on a global step index. Appendix~\ref{app:protocol} tabulates every hyperparameter per arm, and Appendix~\ref{app:repro} gives seeds, selection signals and the rule used to admit a run.
    \item[] Guidelines:
    \begin{itemize}
        \item The answer \answerNA{} means that the paper does not include experiments.
        \item If the paper includes experiments, a \answerNo{} answer to this question will not be perceived well by the reviewers: Making the paper reproducible is important, regardless of whether the code and data are provided or not.
        \item If the contribution is a dataset and\slash or model, the authors should describe the steps taken to make their results reproducible or verifiable. 
        \item Depending on the contribution, reproducibility can be accomplished in various ways. For example, if the contribution is a novel architecture, describing the architecture fully might suffice, or if the contribution is a specific model and empirical evaluation, it may be necessary to either make it possible for others to replicate the model with the same dataset, or provide access to the model. In general. releasing code and data is often one good way to accomplish this, but reproducibility can also be provided via detailed instructions for how to replicate the results, access to a hosted model (e.g., in the case of a large language model), releasing of a model checkpoint, or other means that are appropriate to the research performed.
        \item While NeurIPS does not require releasing code, the conference does require all submissions to provide some reasonable avenue for reproducibility, which may depend on the nature of the contribution. For example
        \begin{enumerate}
            \item If the contribution is primarily a new algorithm, the paper should make it clear how to reproduce that algorithm.
            \item If the contribution is primarily a new model architecture, the paper should describe the architecture clearly and fully.
            \item If the contribution is a new model (e.g., a large language model), then there should either be a way to access this model for reproducing the results or a way to reproduce the model (e.g., with an open-source dataset or instructions for how to construct the dataset).
            \item We recognize that reproducibility may be tricky in some cases, in which case authors are welcome to describe the particular way they provide for reproducibility. In the case of closed-source models, it may be that access to the model is limited in some way (e.g., to registered users), but it should be possible for other researchers to have some path to reproducing or verifying the results.
        \end{enumerate}
    \end{itemize}

\item {\bf Open access to data and code}
    \item[] Question: Does the paper provide open access to the data and code, with sufficient instructions to faithfully reproduce the main experimental results, as described in supplemental material?
    \item[] Answer: \answerYes{} 
    \item[] Justification: The code is available as an anonymized repository at \url{https://anonymous.4open.science/r/BaCP-4721/}. Both datasets are public and standard, the backbones are public torchvision checkpoints, and Appendix F states the protocol and hyperparameters in full. We do not claim bit-for-bit reproduction: the grid was run across several cloud GPUs, and the admission rules of Appendix G rather than a single script define which runs enter the tables.
    \item[] Guidelines:
    \begin{itemize}
        \item The answer \answerNA{} means that paper does not include experiments requiring code.
        \item Please see the NeurIPS code and data submission guidelines (\url{https://neurips.cc/public/guides/CodeSubmissionPolicy}) for more details.
        \item While we encourage the release of code and data, we understand that this might not be possible, so \answerNo{} is an acceptable answer. Papers cannot be rejected simply for not including code, unless this is central to the contribution (e.g., for a new open-source benchmark).
        \item The instructions should contain the exact command and environment needed to run to reproduce the results. See the NeurIPS code and data submission guidelines (\url{https://neurips.cc/public/guides/CodeSubmissionPolicy}) for more details.
        \item The authors should provide instructions on data access and preparation, including how to access the raw data, preprocessed data, intermediate data, and generated data, etc.
        \item The authors should provide scripts to reproduce all experimental results for the new proposed method and baselines. If only a subset of experiments are reproducible, they should state which ones are omitted from the script and why.
        \item At submission time, to preserve anonymity, the authors should release anonymized versions (if applicable).
        \item Providing as much information as possible in supplemental material (appended to the paper) is recommended, but including URLs to data and code is permitted.
    \end{itemize}

\item {\bf Experimental setting/details}
    \item[] Question: Does the paper specify all the training and test details (e.g., data splits, hyperparameters, how they were chosen, type of optimizer) necessary to understand the results?
    \item[] Answer: \answerYes{} 
    \item[] Justification: Section~\ref{sec:objective} gives the optimiser, learning rates, schedule, mask period, epoch budget and batch size, and states the two settings that differ between arms. Appendix~\ref{app:protocol} repeats all of it as a per-arm table. Data splits are stated in Section~\ref{sec:objective}; selection signals, which differ between the contrastive phase and the fine-tunes, are given in Appendix~\ref{app:repro}.
    \item[] Guidelines:
    \begin{itemize}
        \item The answer \answerNA{} means that the paper does not include experiments.
        \item The experimental setting should be presented in the core of the paper to a level of detail that is necessary to appreciate the results and make sense of them.
        \item The full details can be provided either with the code, in appendix, or as supplemental material.
    \end{itemize}

\item {\bf Experiment statistical significance}
    \item[] Question: Does the paper report error bars suitably and correctly defined or other appropriate information about the statistical significance of the experiments?
    \item[] Answer: \answerYes{} 
    \item[] Justification: Every cell of the grid was run at seeds 1, 2 and 3 on both arms, and Appendix~\ref{app:tables} reports each as a mean with a sample standard deviation across those three seeds, a one-sigma bar capturing seed-to-seed variation in initialization, data ordering and augmentation sampling with the data split held fixed. Table~\ref{tab:main} reports the seed means and states the per-cell seed spread, whose median is 0.29 points but which exceeds 0.3 in 65\% of cells; bold marks improvements above $+0.3$. The head and snapshot ablations are single-seed and are labeled as such. No significance test is claimed.
    \item[] Guidelines:
    \begin{itemize}
        \item The answer \answerNA{} means that the paper does not include experiments.
        \item The authors should answer \answerYes{} if the results are accompanied by error bars, confidence intervals, or statistical significance tests, at least for the experiments that support the main claims of the paper.
        \item The factors of variability that the error bars are capturing should be clearly stated (for example, train/test split, initialization, random drawing of some parameter, or overall run with given experimental conditions).
        \item The method for calculating the error bars should be explained (closed form formula, call to a library function, bootstrap, etc.)
        \item The assumptions made should be given (e.g., Normally distributed errors).
        \item It should be clear whether the error bar is the standard deviation or the standard error of the mean.
        \item It is OK to report 1-sigma error bars, but one should state it. The authors should preferably report a 2-sigma error bar than state that they have a 96\% CI, if the hypothesis of Normality of errors is not verified.
        \item For asymmetric distributions, the authors should be careful not to show in tables or figures symmetric error bars that would yield results that are out of range (e.g., negative error rates).
        \item If error bars are reported in tables or plots, the authors should explain in the text how they were calculated and reference the corresponding figures or tables in the text.
    \end{itemize}

\item {\bf Experiments compute resources}
    \item[] Question: For each experiment, does the paper provide sufficient information on the computer resources (type of compute workers, memory, time of execution) needed to reproduce the experiments?
    \item[] Answer: \answerYes{} 
    \item[] Justification: Appendix~\ref{app:compute} names the hardware, a single NVIDIA A100 on a 24-vCPU cloud node, states that runs are serialised with no multi-GPU or multi-node training, and gives the size of the sweep at 540 runs. It also states that the arms are not equal in cost, since a BaCP step forwards four networks against the baseline's one, and that the project consumed compute beyond what is reported, including a discarded earlier generation of runs.
    \item[] Guidelines:
    \begin{itemize}
        \item The answer \answerNA{} means that the paper does not include experiments.
        \item The paper should indicate the type of compute workers CPU or GPU, internal cluster, or cloud provider, including relevant memory and storage.
        \item The paper should provide the amount of compute required for each of the individual experimental runs as well as estimate the total compute. 
        \item The paper should disclose whether the full research project required more compute than the experiments reported in the paper (e.g., preliminary or failed experiments that didn't make it into the paper). 
    \end{itemize}
    
\item {\bf Code of ethics}
    \item[] Question: Does the research conducted in the paper conform, in every respect, with the NeurIPS Code of Ethics \url{https://neurips.cc/public/EthicsGuidelines}?
    \item[] Answer: \answerYes{} 
    \item[] Justification: The research uses two public image-classification benchmarks and public pretrained checkpoints, involves no human subjects and no personal data, and releases no model or dataset. We reviewed the NeurIPS Code of Ethics and find no respect in which the work departs from it.
    \item[] Guidelines:
    \begin{itemize}
        \item The answer \answerNA{} means that the authors have not reviewed the NeurIPS Code of Ethics.
        \item If the authors answer \answerNo, they should explain the special circumstances that require a deviation from the Code of Ethics.
        \item The authors should make sure to preserve anonymity (e.g., if there is a special consideration due to laws or regulations in their jurisdiction).
    \end{itemize}

\item {\bf Broader impacts}
    \item[] Question: Does the paper discuss both potential positive societal impacts and negative societal impacts of the work performed?
    \item[] Answer: \answerYes{} 
    \item[] Justification: Appendix~\ref{app:compute} discusses both directions. The positive case is the standard one for pruning: cheaper storage and inference at a fixed accuracy widens the range of devices a model can run on. On the negative side we note that cheaper inference marginally lowers the cost of deploying any classifier, and, more concretely, that pruning is known to degrade under-represented classes faster than aggregate accuracy suggests, which our aggregate top-1 reporting would not detect.
    \item[] Guidelines:
    \begin{itemize}
        \item The answer \answerNA{} means that there is no societal impact of the work performed.
        \item If the authors answer \answerNA{} or \answerNo, they should explain why their work has no societal impact or why the paper does not address societal impact.
        \item Examples of negative societal impacts include potential malicious or unintended uses (e.g., disinformation, generating fake profiles, surveillance), fairness considerations (e.g., deployment of technologies that could make decisions that unfairly impact specific groups), privacy considerations, and security considerations.
        \item The conference expects that many papers will be foundational research and not tied to particular applications, let alone deployments. However, if there is a direct path to any negative applications, the authors should point it out. For example, it is legitimate to point out that an improvement in the quality of generative models could be used to generate Deepfakes for disinformation. On the other hand, it is not needed to point out that a generic algorithm for optimizing neural networks could enable people to train models that generate Deepfakes faster.
        \item The authors should consider possible harms that could arise when the technology is being used as intended and functioning correctly, harms that could arise when the technology is being used as intended but gives incorrect results, and harms following from (intentional or unintentional) misuse of the technology.
        \item If there are negative societal impacts, the authors could also discuss possible mitigation strategies (e.g., gated release of models, providing defenses in addition to attacks, mechanisms for monitoring misuse, mechanisms to monitor how a system learns from feedback over time, improving the efficiency and accessibility of ML).
    \end{itemize}
    
\item {\bf Safeguards}
    \item[] Question: Does the paper describe safeguards that have been put in place for responsible release of data or models that have a high risk for misuse (e.g., pre-trained language models, image generators, or scraped datasets)?
    \item[] Answer: \answerNA{} 
    \item[] Justification: No model, dataset or other artefact is released with this paper, and the work poses no dual-use or misuse risk requiring safeguards. The datasets used are standard public image-classification benchmarks.
    \item[] Guidelines:
    \begin{itemize}
        \item The answer \answerNA{} means that the paper poses no such risks.
        \item Released models that have a high risk for misuse or dual-use should be released with necessary safeguards to allow for controlled use of the model, for example by requiring that users adhere to usage guidelines or restrictions to access the model or implementing safety filters. 
        \item Datasets that have been scraped from the Internet could pose safety risks. The authors should describe how they avoided releasing unsafe images.
        \item We recognize that providing effective safeguards is challenging, and many papers do not require this, but we encourage authors to take this into account and make a best faith effort.
    \end{itemize}

\item {\bf Licenses for existing assets}
    \item[] Question: Are the creators or original owners of assets (e.g., code, data, models), used in the paper, properly credited and are the license and terms of use explicitly mentioned and properly respected?
    \item[] Answer: \answerYes{} 
    \item[] Justification: Appendix~\ref{app:compute} credits and licenses every asset used: CIFAR-10 and CIFAR-100 \citep{krizhevsky2009learning} under their standard research release, the torchvision ImageNet-pretrained backbones \citep{he2016deep,simonyan2014very,sandler2018mobilenetv2} and PyTorch itself under BSD-3-Clause. The three pruning criteria \citep{han2015learning,lee2018snip,sun2023simple} are cited and were reimplemented from their published descriptions rather than taken from the authors' code.
    \item[] Guidelines:
    \begin{itemize}
        \item The answer \answerNA{} means that the paper does not use existing assets.
        \item The authors should cite the original paper that produced the code package or dataset.
        \item The authors should state which version of the asset is used and, if possible, include a URL.
        \item The name of the license (e.g., CC-BY 4.0) should be included for each asset.
        \item For scraped data from a particular source (e.g., website), the copyright and terms of service of that source should be provided.
        \item If assets are released, the license, copyright information, and terms of use in the package should be provided. For popular datasets, \url{paperswithcode.com/datasets} has curated licenses for some datasets. Their licensing guide can help determine the license of a dataset.
        \item For existing datasets that are re-packaged, both the original license and the license of the derived asset (if it has changed) should be provided.
        \item If this information is not available online, the authors are encouraged to reach out to the asset's creators.
    \end{itemize}

\item {\bf New assets}
    \item[] Question: Are new assets introduced in the paper well documented and is the documentation provided alongside the assets?
    \item[] Answer: \answerNA{} 
    \item[] Justification: The paper releases no new dataset, model, benchmark or other asset. It reports a controlled study of an existing objective on public data and public pretrained checkpoints.
    \item[] Guidelines:
    \begin{itemize}
        \item The answer \answerNA{} means that the paper does not release new assets.
        \item Researchers should communicate the details of the dataset\slash code\slash model as part of their submissions via structured templates. This includes details about training, license, limitations, etc. 
        \item The paper should discuss whether and how consent was obtained from people whose asset is used.
        \item At submission time, remember to anonymize your assets (if applicable). You can either create an anonymized URL or include an anonymized zip file.
    \end{itemize}

\item {\bf Crowdsourcing and research with human subjects}
    \item[] Question: For crowdsourcing experiments and research with human subjects, does the paper include the full text of instructions given to participants and screenshots, if applicable, as well as details about compensation (if any)? 
    \item[] Answer: \answerNA{} 
    \item[] Justification: The paper involves no crowdsourcing and no research with human subjects. All experiments are training runs on public image-classification benchmarks.
    \item[] Guidelines:
    \begin{itemize}
        \item The answer \answerNA{} means that the paper does not involve crowdsourcing nor research with human subjects.
        \item Including this information in the supplemental material is fine, but if the main contribution of the paper involves human subjects, then as much detail as possible should be included in the main paper. 
        \item According to the NeurIPS Code of Ethics, workers involved in data collection, curation, or other labor should be paid at least the minimum wage in the country of the data collector. 
    \end{itemize}

\item {\bf Institutional review board (IRB) approvals or equivalent for research with human subjects}
    \item[] Question: Does the paper describe potential risks incurred by study participants, whether such risks were disclosed to the subjects, and whether Institutional Review Board (IRB) approvals (or an equivalent approval/review based on the requirements of your country or institution) were obtained?
    \item[] Answer: \answerNA{} 
    \item[] Justification: The paper involves no research with human subjects, so no IRB review or equivalent was required.
    \item[] Guidelines:
    \begin{itemize}
        \item The answer \answerNA{} means that the paper does not involve crowdsourcing nor research with human subjects.
        \item Depending on the country in which research is conducted, IRB approval (or equivalent) may be required for any human subjects research. If you obtained IRB approval, you should clearly state this in the paper. 
        \item We recognize that the procedures for this may vary significantly between institutions and locations, and we expect authors to adhere to the NeurIPS Code of Ethics and the guidelines for their institution. 
        \item For initial submissions, do not include any information that would break anonymity (if applicable), such as the institution conducting the review.
    \end{itemize}

\item {\bf Declaration of LLM usage}
    \item[] Question: Does the paper describe the usage of LLMs if it is an important, original, or non-standard component of the core methods in this research? Note that if the LLM is used only for writing, editing, or formatting purposes and does \emph{not} impact the core methodology, scientific rigor, or originality of the research, declaration is not required.
    \item[] Answer: \answerNA{} 
    \item[] Justification: No LLM is a component of the method, the experiments or the analysis. BaCP is a training objective for convolutional image classifiers, and every result reported here comes from those training runs.
    \item[] Guidelines:
    \begin{itemize}
        \item The answer \answerNA{} means that the core method development in this research does not involve LLMs as any important, original, or non-standard components.
        \item Please refer to our LLM policy in the NeurIPS handbook for what should or should not be described.
    \end{itemize}

\end{enumerate}

\end{document}